\documentclass[12pt]{article}

\pdfoutput=1 

\usepackage{amsbsy, amstext, amssymb, amsthm, amsmath,bm}
\usepackage{graphicx,booktabs,natbib, hyperref}
\usepackage[lined,boxed,linesnumbered]{algorithm2e}
\usepackage[margin=1in]{geometry}
\usepackage{enumerate}
\usepackage{subfig, footmisc}

\hypersetup{
	colorlinks = true, % use color instead of boxes
	urlcolor = blue, % external (email/http) links
	linkcolor = blue, % internal links (figures/tables/equations)
	citecolor = blue, % citations
}

\newcommand{\E}{\mathrm{E}}

\newcommand{\rank}{\mathrm{rank}}

\newcommand{\Abf}{{\bm A}}

\newcommand{\Cbf}{{\bm C}}
\newcommand{\Dbf}{{\bm D}}

\newcommand{\Fbf}{{\bm F}}

\newcommand{\Ibf}{{\bm I}}

\newcommand{\Mbf}{{\bm M}}

\newcommand{\Pbf}{{\bm P}}
\newcommand{\Qbf}{{\bm Q}}
\newcommand{\Rbf}{{\bm R}}

\newcommand{\Ubf}{{\bm U}}
\newcommand{\Vbf}{{\bm V}}

\newcommand{\Xbf}{{\bm X}}

\newcommand{\Zbf}{{\bm Z}}

\newcommand{\bbf}{{\bm b}}
\newcommand{\cbf}{{\bm c}}
\newcommand{\dbf}{{\bm d}}
\newcommand{\ebf}{{\bm e}}

\newcommand{\rbf}{{\bm r}}
\newcommand{\sbf}{{\bm s}}
\newcommand{\ubf}{{\bm u}}
\newcommand{\vbf}{{\bm v}}

\newcommand{\xbf}{{\bm x}}
\newcommand{\ybf}{{\bm y}}
\newcommand{\zbf}{{\bm z}}

\newcommand{\zerobf}{{\mathbf 0}}
\newcommand{\onebf}{{\mathbf 1}}

\newcommand{\greekbold}[1]{\mbox{\boldmath $#1$}}
\newcommand{\alphabf}{\greekbold{\alpha}}
\newcommand{\betabf}{\greekbold{\beta}}

\newcommand{\gammabf}{\greekbold{\gamma}}
\newcommand{\lambdabf}{\greekbold{\lambda}}
\newcommand{\mubf}{\greekbold{\mu}}
\newcommand{\nubf}{\greekbold{\nu}}

\newcommand{\thetabf}{\greekbold{\theta}}
\newcommand{\varepsilonbf}{\greekbold{\varepsilon}}

\newcommand{\Sigmabf}{\greekbold{\Sigma}}

\newcommand{\reals}{\mathbb{R}}
\DeclareMathOperator*{\argmin}{arg\,min}

\title{Algorithms for Fitting the Constrained Lasso}

\author{
Brian R. Gaines and Hua Zhou\footnote{Brian R. Gaines, Department of Statistics, North Carolina State University, Raleigh, NC 27695 (E-mail: \href{mailto:brgaines@ncsu.edu}{brgaines@ncsu.edu}).  Hua Zhou, Department of Biostatistics, University of California, Los Angeles (UCLA), Los Angeles, CA 90095 (E-mail: \href{mailto:huazhou@ucla.edu}{huazhou@ucla.edu}).}
}
\date{} 
\begin{document}
\maketitle
\begin{abstract}
We compare alternative computing strategies for solving the constrained lasso problem.  As its name suggests, the constrained lasso extends the widely-used lasso to handle linear constraints, which allow the user to incorporate prior information into the model.  In addition to quadratic programming, we employ the alternating direction method of multipliers (ADMM) and also derive an efficient solution path algorithm.  Through both simulations and real data examples, we compare the different algorithms and provide practical recommendations in terms of efficiency and accuracy for various sizes of data.  We also show that, for an arbitrary penalty matrix, the generalized lasso can be transformed to a constrained lasso, while the converse is not true.  Thus, our methods can also be used for estimating a generalized lasso, which has wide-ranging applications.  Code for implementing the algorithms is freely available in the \textsc{Matlab} toolbox \href{https://github.com/Hua-Zhou/SparseReg}{\tt SparseReg}.  
\end{abstract}


\section{Introduction}
\label{sec:intro}
\begin{figure}
	\begin{center}
		\includegraphics[scale=0.8]{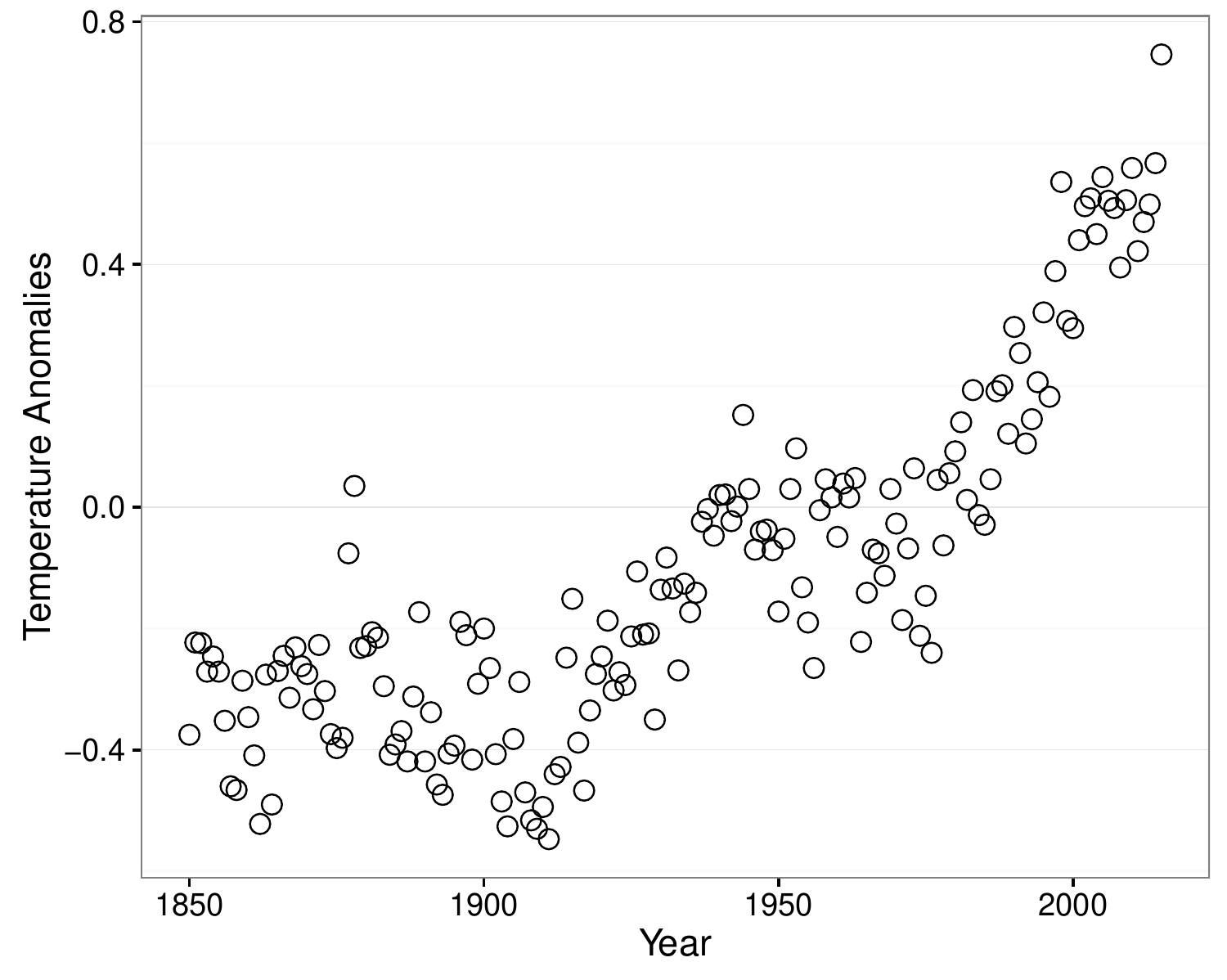} % exported as 6x4.8'' pdf from R
	\caption{Global Warming Data. \small Annual temperature anomalies relative to the 1961-1990 average.}
	\label{fig:warmingData}
	\end{center}
\end{figure}

Our focus is on estimating the constrained lasso problem \citep{JamesPaulsonRusmevichientong14CLasso} 
\begin{eqnarray}
	&\text{minimize}& \quad \frac 12 \| \ybf - \Xbf \betabf\|_2^2 + \rho \|\betabf\|_1 \label{eqn:LS-constrlasso} \\
	&\text{subject to}& \quad \Abf \betabf = \bbf \text{ and } \Cbf \betabf \le \dbf,	\nonumber
\end{eqnarray}
where $\ybf \in \mathbb{R}^n$ is the response vector, $\Xbf \in \mathbb{R}^{n \times p}$ is the design matrix of predictors or covariates, $\betabf \in \mathbb{R}^p$ is the vector of unknown regression coefficients, and $\rho \ge 0$ is a tuning parameter that controls the amount of regularization.  It is assumed that the constraint matrices, $\Abf$ and $\Cbf$, both have full row rank.  As its name suggests, the constrained lasso augments the standard lasso \citep{tibs96}  with linear equality and inequality constraints.  While the use of the $\ell_1$ penalty allows a user to impose prior knowledge on the coefficient estimates in terms of sparsity, the constraints provide an additional vehicle for prior knowledge to be incorporated into the solution.  For example, consider the annual data on temperature anomalies given in Figure~\ref{fig:warmingData}.  As has been previously noted in the literature on isotonic regression, in general temperature appears to increase monotonically over the time period of 1850 to 2015 \citep{wu2001isotonic, Tibshirani2011Isotonic}.  This monotonicity can be imposed on the coefficient estimates using the constrained lasso with the inequality constraint matrix 
\begin{eqnarray*}
	\Cbf = \begin{pmatrix} 
	1 & -1 \\
	& 1 & -1 \\
	& & \ddots & \ddots \\
	& & & & 1 & -1 
	\end{pmatrix}
\end{eqnarray*}
and $\dbf = \zerobf$.  The lasso with a monotonic ordering of the coefficients was referred to by \citet{Suo2016ordered} as the \textit{ordered lasso}, and is a special case of the constrained lasso \eqref{eqn:LS-constrlasso}.

Another example of the constrained lasso that has appeared in the literature is the \textit{positive lasso}.  First mentioned in the seminal work of \citet{efron2004lars}, the positive lasso requires the lasso coefficients to be non-negative.  This variant of the lasso has seen applications in areas such as vaccine design \citep{hu2015vaccine}, nuclear material detection \citep{kump2012rival}, document classification \citep{elarini2013documents}, and portfolio management \citep{wu2014nonnegative}.  The positive lasso is a special case of the constrained lasso \eqref{eqn:LS-constrlasso} with $\Cbf = -\Ibf_p$ and $\dbf = \zerobf_p$.  Additionally, there are several other examples throughout the literature where the original lasso is augmented with additional information in the form of linear equality or inequality constraints.  \citet{huang2013proteinlasso} constrained the lasso estimates to be in the unit interval to interpret the coefficients as probabilities associated with the presence of a certain protein in a cell or tissue.  The lasso with a sum-to-zero constraint on the coefficients has been used for regression \citep{shi2016microbiome} and variable selection \citep{lin2014varselectcompositional} with compositional data as covariates.  Compositional data are multivariate data that represent proportions of a whole and thus must sum to one, and arrive in applications such as consumer spending in economics, topic consumption of documents in machine learning, and the human microbiome \citep{lin2014varselectcompositional}.  Lastly, simplex constraints were utilized by \citet{huang2013brain} when using the lasso to estimate edge weights in brain networks.  Thus, the constrained lasso is a very flexible framework for imposing additional knowledge and structure onto the lasso coefficient estimates.  

During the preparation of our manuscript, we became aware of unpublished work by \citet{he2011} that also derived a solution path algorithm for solving the constrained lasso.  However, our approach to deriving the path algorithm is completely different and is more in line with the literature on solution path algorithms \citep{rosset2007piecewise}, especially in the presence of constraints \citep{ZhouLange2013Constrained}.  Additionally, we address how our algorithms can be adapted to work in the high dimensional setting where $n < p$, which was not done by \citet{he2011}.  Furthermore, the approach by \citet{he2011} decomposes the parameter vector, $\betabf$, into its positive and negative parts, $\betabf = \betabf^+ - \betabf^-$, thus doubling the size of the problem.  On the other hand, we work directly with the original coefficient vector at the benefit of computational efficiency and notational simplicity.  Lastly, another important contribution of our work is the implementation of our algorithms in the \href{https://github.com/Hua-Zhou/SparseReg}{\tt SparseReg} \textsc{Matlab} toolbox available on Github.  

The constrained lasso was also studied by \citet{JamesPaulsonRusmevichientong14CLasso} in an earlier version of their manuscript on penalized and constrained (PAC) regression.  The current PAC regression framework extends \eqref{eqn:LS-constrlasso} by using a negative log likelihood for the loss function to also cover generalized linear models (GLMs), and thus is more general than the problem we study.  However, the increased generality of their method comes at the cost of computational efficiency.  Furthermore, their path algorithm is not a traditional solution path algorithm as it is fit on a pre-specified grid of tuning parameters, which is fundamentally different from our path following strategy.  Additionally, we believe the squared-error loss function merits additional attention given its widespread use with the $\ell_1$ penalty, and also since the constrained lasso is a natural approach to solving constrained least squares problems in the increasingly common high-dimensional setting.  \citet{Hu2015} studied the constrained generalized lasso, which reduces to the constrained lasso when no penalty matrix is included ($\Dbf = \Ibf_p$).  However, they do not derive a solution path algorithm but instead develop a coordinate descent algorithm for fixed values of the tuning parameter.  

The rest of the article is organized as follows.  In Section~\ref{sec:genlasso2classo}, we demonstrate a new connection between the constrained lasso and the generalized lasso, which shows that the latter can always be transformed and solved as a constrained lasso, even when the penalty matrix is rank deficient.  Given the flexibility of the generalized lasso, this result greatly extends the applicability of our algorithms and results.  Various algorithms to solve the constrained lasso, including quadratic programming (QP), the alternating direction method of multipliers (ADMM), and a novel path following algorithm, are derived in Section~\ref{sec:algo}.  Degrees of freedom are established in Section~\ref{sec:df} before simulation results that compare the performance of the various algorithms are presented in Section~\ref{sec:sims}.  The main result from the simulations is that, in terms of run time, the solution path algorithm is more efficient than the other approaches when the coefficient estimates at more than a handful of values of the tuning parameter are desired.  Real data examples that highlight the flexibility of the constrained lasso are given in Section~\ref{sec:realData}, while Section~\ref{sec:conclusion} concludes.  

\section{Connection to the Generalized Lasso}
\label{sec:genlasso2classo}
Another flexible lasso formulation is the generalized lasso \citep{TibshiraniTaylor11GenLasso}
\begin{eqnarray} 
	&\text{minimize}& \quad \frac 12 \| \ybf - \Xbf \betabf\|_2^2 + \rho \|\Dbf \betabf\|_1, \label{eqn:LS-genlasso}
\end{eqnarray}
where $\Dbf \in \mathbb{R}^{m \times p}$ is a fixed, user-specified regularization matrix.  Certain choices of $\Dbf$ correspond to  different versions of the lasso, including the original lasso, various forms of the fused lasso, and trend filtering.  It has been observed that \eqref{eqn:LS-genlasso} can be transformed to a standard lasso when $\Dbf$ has full row rank \citep{TibshiraniTaylor11GenLasso}, while it can be transformed to a constrained lasso when $\Dbf$ has full column rank \citep{JamesPaulsonRusmevichientong14CLasso}.  Here we demonstrate that a generalized lasso \eqref{eqn:LS-genlasso} can be transformed to a constrained lasso \eqref{eqn:LS-constrlasso} for an arbitrary penalty matrix $\Dbf$.

Assume that $\rank(\Dbf) = r$, and consider the singular value decomposition (SVD)
\begin{eqnarray*}
	\Dbf = \Ubf \Sigmabf \Vbf^T = \begin{pmatrix} \Ubf_1, \Ubf_2 \end{pmatrix} \begin{pmatrix} \Sigmabf_1 & \zerobf \\ \zerobf & \zerobf \end{pmatrix} \begin{pmatrix} \Vbf_1^T \\ \Vbf_2^T \end{pmatrix} = \Ubf_1 \Sigmabf_1 \Vbf_1^T,
\end{eqnarray*}
where $\Ubf_1 \in \mathbb{R}^{m \times r}$, $\Ubf_2 \in \mathbb{R}^{m \times (m-r)}$, $\Sigmabf_1 \in \mathbb{R}^{r \times r}$, $\Vbf_1 \in \mathbb{R}^{p \times r}$, and $\Vbf_2 \in \mathbb{R}^{p \times (p-r)}$. We define an augmented matrix 
$$
	\tilde \Dbf = \begin{pmatrix} \Ubf_1 \Sigmabf_1 \Vbf_1^T \\ \Vbf_2^T \end{pmatrix} = \begin{pmatrix} \Ubf_1 \Sigmabf_1 & \zerobf \\ \zerobf & \Ibf_{p-r} \end{pmatrix} \begin{pmatrix} \Vbf_1^T \\ \Vbf_2^T \end{pmatrix} = \begin{pmatrix} \Ubf_1 \Sigmabf_1 & \zerobf \\ \zerobf & \Ibf_{p-r} \end{pmatrix} \Vbf^T
$$
and use the following change of variables
\begin{eqnarray}
	\begin{pmatrix} \alphabf \\ \gammabf \end{pmatrix} = \tilde \Dbf \betabf = \begin{pmatrix} \Ubf_1 \Sigmabf_1 \Vbf_1^T \\ \Vbf_2^T \end{pmatrix} \betabf,	\label{eqn:transformation}
\end{eqnarray}
where $\alphabf \in \mathbb{R}^m$ and $\gammabf \in \mathbb{R}^{p-r}$.  Since the matrix $\Vbf_2^T$ forms a basis for the nullspace of $\Dbf$, $\mathcal{N}(\Dbf)$, it has rank $p - r$ and its columns are linearly independent of the columns of $\Dbf$.  Thus, the augmented matrix $\tilde \Dbf$ has full column rank, and the new variables $\begin{pmatrix} \alphabf \\ \gammabf \end{pmatrix}$ uniquely determine $\betabf$ via
\begin{eqnarray*}
	\betabf &=& (\tilde \Dbf^T \tilde \Dbf)^{-1} \tilde \Dbf^T \begin{pmatrix} \alphabf \\ \gammabf \end{pmatrix} \\
	&=& (\tilde \Dbf^T \tilde \Dbf)^{-1} \Vbf_1 \Sigmabf_1 \Ubf_1^T  \alphabf +  (\tilde \Dbf^T \tilde \Dbf)^{-1} \Vbf_2  \gammabf \\
	&=& \Vbf_1 \Sigmabf_1^{-1} \Ubf_1^T  \alphabf + \Vbf_2  \gammabf \\
	&=& \Dbf^+ \alphabf + \Vbf_2 \gammabf, 
\end{eqnarray*}
where $\Dbf^+$ denotes the Moore-Penrose inverse of the matrix $\Dbf$.  However, since the original change of variables is $\begin{pmatrix} \alphabf \\ \gammabf \end{pmatrix} = \tilde \Dbf \betabf$, $\betabf$ is uniquely determined if and only if
\begin{eqnarray*}
	\begin{pmatrix} \alphabf \\ \gammabf \end{pmatrix} \in {\cal C}(\tilde \Dbf) = {\cal C} \left( \begin{pmatrix} \Ubf_1 \Sigmabf_1 & \zerobf \\ \zerobf & \Ibf_{p-r} \end{pmatrix} \right),
\end{eqnarray*}
if and only if
\begin{eqnarray*}
	\alphabf \in {\cal C}(\Ubf_1 \Sigmabf_1) = {\cal C}(\Ubf_1) = {\cal C}(\Dbf),
\end{eqnarray*}
if and only if
\begin{eqnarray*}
	\Ubf_2^T \alphabf = \zerobf_{m-r}, 
\end{eqnarray*}
where ${\cal C}(\Dbf)$ is the column space of the matrix $\Dbf$.  
Therefore, the generalized lasso problem \eqref{eqn:LS-genlasso} is equivalent to a constrained lasso problem
\begin{eqnarray}
	&\text{minimize}& \quad \frac 12 \left\| \ybf -  \Xbf \Dbf^+  \alphabf -  \Xbf \Vbf_2  \gammabf \right\|_2^2 + \rho \|\alphabf\|_1 \label{eqn:genlasso-in-constrlasso-form} \\
	&\text{subject to}& \quad \Ubf_2^T \alphabf = \zerobf,	\nonumber
\end{eqnarray}
where $\gammabf$ remains unpenalized.  There are three special cases of interest:  
\begin{enumerate}
\item When $\Dbf$ has full row rank, $r=m$, the matrix $\Ubf_2$ is null and the constraint $\Ubf_2^T \alphabf = \zerobf$ vanishes, reducing to a standard lasso as observed by \citet{TibshiraniTaylor11GenLasso}. 

\item When $\Dbf$ has full column rank, $r=p$, the matrix $\Vbf_2$ is null and the term $\Xbf \Vbf_2 \gammabf$ drops, resulting in a constrained lasso as observed by \citet{JamesPaulsonRusmevichientong14CLasso}. 
 
\item When $\Dbf$ does not have full rank, $r < \min(m,p)$, the above problem \eqref{eqn:genlasso-in-constrlasso-form} can be simplified to a constrained lasso problem only in $\alphabf$ by noticing that minimizing \eqref{eqn:genlasso-in-constrlasso-form} with respect to $\gammabf$ yields
\begin{eqnarray*}
	\Xbf \Vbf_2 \hat \gammabf &=& \Pbf_{\Xbf \Vbf_2} ( \ybf -  \Xbf \Dbf^+  \alphabf)
\end{eqnarray*}
for any $\alphabf$, where $\Pbf_{\Xbf \Vbf_2}$ is the orthogonal projection onto the column space ${\cal C}(\Xbf \Vbf_2)$. Thus, for an arbitrary penalty matrix $\Dbf$, using the change of variables \eqref{eqn:transformation} we end up with a constrained lasso problem
\begin{eqnarray*}
	&\text{minimize}& \quad \frac 12 \| \tilde{\ybf} - \tilde{\Xbf}  \alphabf \|_2^2 + \rho \|\alphabf\|_1 \\
	&\text{subject to}& \quad \Ubf_2^T \alphabf = \zerobf,
\end{eqnarray*}
where $\tilde{\ybf} = (\Ibf - \Pbf_{\Xbf \Vbf_2}) \ybf$ and $\tilde{\Xbf} = (\Ibf - \Pbf_{\Xbf \Vbf_2})  \Xbf \Dbf^+$.  The solution path $\hat \alphabf(\rho)$ can be translated back to that of the original generalized lasso problem via the affine transform
\begin{eqnarray*}
	\hat \betabf(\rho) 
	&=& \Vbf_1 \Sigmabf_1^{-1} \Ubf_1^T  \hat \alphabf (\rho) + \Vbf_2 (\Vbf_2^T \Xbf^T \Xbf \Vbf_2)^- \Vbf_2^T \Xbf^T [\ybf - \Xbf \Dbf^+ \hat\alphabf(\rho)] \\
	&=&  [\Ibf - \Vbf_2 (\Vbf_2^T \Xbf^T \Xbf \Vbf_2)^- \Vbf_2^T \Xbf^T \Xbf] \Dbf^+ \hat\alphabf(\rho) \\
	& & \quad + \Vbf_2 (\Vbf_2^T \Xbf^T \Xbf \Vbf_2)^- \Vbf_2^T \Xbf^T \ybf,  
\end{eqnarray*}
where $\Xbf^-$ denotes the generalized inverse of a matrix $\Xbf$.
\end{enumerate}
Thus, any generalized lasso problem can be reformulated as a constrained lasso, so the algorithms and results presented here are applicable to a large class of problems.  However, it is not always possible to transform a constrained lasso into a generalized lasso, as detailed in Appendix~\ref{sec:classo2genlasso}.

\section{Algorithms}
\label{sec:algo}

In this section, we derive three different algorithms for estimating the constrained lasso \eqref{eqn:LS-constrlasso}.  Throughout this section, we assume that $\Xbf$ has full column rank, which necessitates that $n > p$.  For the increasingly prevalent high-dimensional case where $n < p$, we follow the standard approach in the related literature \citep{TibshiraniTaylor11GenLasso, Hu2015, arnold2016efficient} and add a small ridge penalty, $\frac{\varepsilon}{2} \| \betabf \|_2^2$, to the original objective function in \eqref{eqn:LS-constrlasso}, where $\varepsilon$ is some small constant, such as $10^{-4}$.  The problem then becomes
\begin{eqnarray}
	&\text{minimize}& \quad \frac 12 \| \ybf - \Xbf \betabf\|_2^2 + \rho \|\betabf\|_1 + \frac{\varepsilon}{2} \| \betabf \|_2^2 \label{eqn:LS-constrlassoRidge} \\
	&\text{subject to}& \quad \Abf \betabf = \bbf \text{ and } \Cbf \betabf \le \dbf.	\nonumber
\end{eqnarray}
Note that the objective \eqref{eqn:LS-constrlassoRidge} can be re-arranged into standard constrained lasso form \eqref{eqn:LS-constrlasso}

\begin{eqnarray}
	&\text{minimize}& \quad \frac 12 \| \ybf^* - (\Xbf^*) \betabf\|_2^2 + \rho \|\betabf\|_1 \label{eqn:LS-constrlassoAugmented} \\
	&\text{subject to}& \quad \Abf \betabf = \bbf \text{ and } \Cbf \betabf \le \dbf,	\nonumber
\end{eqnarray}
using the augmented data $\ybf^* = \begin{pmatrix} \ybf \\ \zerobf \end{pmatrix}$ and $\Xbf^* = \begin{pmatrix} \Xbf \\ \sqrt{\varepsilon} \Ibf_p \end{pmatrix} $.  The augmented design matrix has full column rank, so the following algorithms can then be applied to the augmented form \eqref{eqn:LS-constrlassoAugmented}.  As discussed by \citet{TibshiraniTaylor11GenLasso}, this approach is attractive for more than just computational reasons, as the inclusion of the ridge penalty may also improve predictive accuracy.  
 
Before deriving the algorithms, we first define some notation.  For a vector $\vbf$ and index set $\mathcal{S}$, let $\vbf_{\mathcal{S}}$ be the sub-vector of size $| \mathcal{S} |$ containing the elements of $\vbf$ corresponding to the indices in $\mathcal{S}$, where $| \cdot |$ denotes the cardinality or size of the index set.  Similarly, for a matrix $\Mbf$ and another index set ${\mathcal{T}}$, the matrix $\Mbf_{\mathcal{S}, \mathcal{T}}$ contains the rows from $\Mbf$ corresponding to the indices in $\mathcal{S}$ and the columns of $\Mbf$ from the indices in $\mathcal{T}$.  We use a colon, :, when all indices along one of the dimensions are included.  That is, $\Mbf_{\mathcal{S}, :}$ contains the rows from $\Mbf$ corresponding to $\mathcal{S}$ but all of the columns in $\Mbf$.
 
%% QP %%
\subsection{Quadratic Programming}

Our first approach is to use quadratic programming to solve the constrained lasso problem \eqref{eqn:LS-constrlasso}.  The key is to decompose $\betabf$ into its positive and negative parts, $\betabf = \betabf^+ - \betabf^-$, as the relation $| \betabf | = \betabf^+ + \betabf^-$ handles the $\ell_1$ penalty term.  By plugging these into \eqref{eqn:LS-constrlasso} and adding the additional non-negativity constraints on $\betabf^+$ and $ \betabf^-$, the constrained lasso is formulated as a standard quadratic program of $2p$ variables, 
\begin{eqnarray}
	&\text{minimize}& \quad \frac 12 \begin{pmatrix} \betabf^+ \\ \betabf^- \end{pmatrix}^T \begin{pmatrix} \Xbf^T \Xbf & - \Xbf^T \Xbf \\ -\Xbf^T \Xbf & \Xbf^T \Xbf \end{pmatrix} \begin{pmatrix} \betabf^+ \\ \betabf^- \end{pmatrix} + \left( \rho \onebf_{2p} - \begin{pmatrix} \Xbf^T \ybf \\ -\Xbf^T \ybf \end{pmatrix} \right)^T \begin{pmatrix} \betabf^+ \\ \betabf^- \end{pmatrix} \label{eqn:LS-constrlasso-QP} \\
	&\text{subject to}& \quad \begin{pmatrix} \Abf & -\Abf \end{pmatrix} \begin{pmatrix} \betabf^+ \\ \betabf^- \end{pmatrix} = \bbf \nonumber \\
	& & \quad \begin{pmatrix} \Cbf & -\Cbf \end{pmatrix} \begin{pmatrix} \betabf^+ \\ \betabf^- \end{pmatrix} \le \dbf	\nonumber \\
	& & \quad \betabf^+ \ge \zerobf, \quad \betabf^- \ge \zerobf. \nonumber
\end{eqnarray}
\textsc{Matlab}'s {\tt quadprog} function is able to scale up to $p \sim 10^2\text{-}10^3$, while the commercial {\sc Gurobi Optimizer} is able to scale up to $p \sim 10^3\text{-}10^4$.

%% ADMM %%
\subsection{ADMM}
\label{sec:ADMM}
The next algorithm we apply to the constrained lasso problem \eqref{eqn:LS-constrlasso} is the alternating direction method of multipliers (ADMM).  The ADMM algorithm has experienced renewed interest in statistics and machine learning applications in recent years; see \citet{Boyd11ADMM} for a recent survey.  In general ADMM is an algorithm to solve a problem that features a separable objective but coupling constraints,
\begin{eqnarray*}
	&\text{minimize}& \quad f(\xbf) + g(\zbf) \\
	&\text{subject to}& \Mbf \xbf + \Fbf \zbf = \cbf,  
\end{eqnarray*}
where $f, g: \mathbb{R}^{p} \mapsto \mathbb{R} \cup \{\infty\}$ are closed proper convex functions.  The idea is to employ block coordinate descent to the augmented Lagrangian function followed by an update of the dual variables $\nubf$, 
\begin{eqnarray*}
	\xbf^{(t+1)} &\gets& \argmin_{\xbf}    \mathcal{L}_{\tau}(\xbf, \zbf^{(t)}, \nubf^{(t)}) 	\\
	\zbf^{(t+1)} &\gets&  \argmin_{\zbf}    \mathcal{L}_{\tau}(\xbf^{(t + 1)}, \zbf, \nubf^{(t)}) 	\\
	\nubf^{(t+1)} &\gets& \nubf^{(t)} + \tau(\Mbf \xbf^{(t+1)} + \Fbf \zbf^{(t+1)} - \cbf),
\end{eqnarray*}
where $t$ is the iteration counter and the augmented Lagrangian is 
\begin{eqnarray}
	\mathcal{L}_{\tau}(\xbf, \zbf, \nubf) = f(\xbf) + g(\zbf) + \nubf^T(\Mbf \xbf + \Fbf \zbf - \cbf) + \frac{\tau}{2} \| \Mbf \xbf + \Fbf \zbf - \cbf \|^2_2. \label{eq:augLagrange}
\end{eqnarray}
Often it is more convenient to work with the equivalent scaled form of ADMM, which scales the dual variable and combines the linear and quadratic terms in the augmented Lagrangian \eqref{eq:augLagrange}.  The updates become
\begin{eqnarray*}
	\xbf^{(t+1)} &\gets& \argmin_{\xbf}    f(\xbf) + \frac{\tau}{2} \| \Mbf \xbf + \Fbf \zbf^{(t)} - \cbf + \ubf^{(t)}   \|^2_2	\\
	\zbf^{(t+1)} &\gets&  \argmin_{\zbf}   g(\zbf) +  \frac{\tau}{2} \| \Mbf \xbf^{(t+1)} + \Fbf \zbf - \cbf + \ubf^{(t)} \|^2_2  	\\
	\ubf^{(t+1)} &\gets& \ubf^{(t)} + \Mbf \xbf^{(t+1)} + \Fbf \zbf^{(t+1)} - \cbf,
\end{eqnarray*}
where $\ubf = \nubf/\tau $  is the scaled dual variable.  The scaled form is especially useful in the case where $\Mbf = \Fbf = \Ibf$, as the updates can be rewritten as \begin{eqnarray*}
	\xbf^{(t+1)} &\gets& \mathbf{prox}_{\tau f} (\zbf^{(t)} - \cbf + \ubf^{(t)})	\\
	\zbf^{(t+1)} &\gets& \mathbf{prox}_{\tau g} (\xbf^{(t+1)} - \cbf +  \ubf^{(t)})	\\
	\ubf^{(t+1)} &\gets& \ubf^{(t)} + \xbf^{(t+1)} + \zbf^{(t+1)} - \cbf,
\end{eqnarray*}
where $\mathbf{prox}_{\tau f}$ is the proximal mapping of a function $f$ with parameter $\tau > 0$.  Recall that the proximal mapping is defined as
\begin{eqnarray*}
	\mathbf{prox}_{\tau f} (\vbf) = \underset{\xbf}{\text{argmin}} \left( f(\xbf) + \frac{1}{2\tau} \|\xbf - \vbf\|_2^2 \right).
\end{eqnarray*}
One benefit of using the scaled form for ADMM is that, in many situations including the constrained lasso, the proximal mappings have simple, closed form solutions, resulting in straightforward ADMM updates.  To apply ADMM to the constrained lasso, we identify $f$ as the objective in \eqref{eqn:LS-constrlasso} and $g$ as the indicator function of the constraint set ${\cal C} = \{\betabf \in \mathbb{R}^p: \Abf \betabf = \bbf, \Cbf \betabf \le \dbf\}$,
\begin{eqnarray*}
	g(\betabf) = \chi_{\cal C} = \begin{cases} \infty & \betabf \notin {\cal C} \\ 0 & \betabf \in {\cal C}. \end{cases}
\end{eqnarray*}
For the updates, $\text{prox}_{\tau f}$ is a regular lasso problem and $\text{prox}_{\tau g}$ is a projection onto the affine space ${\cal C}$ (Algorithm~\ref{algo:ADMM-constrlasso}). The projection onto ${\cal C}$ is more efficiently solved by the dual problem, which has a smaller number of variables.

\begin{algorithm}[t]
Initialize $\betabf^{(0)} = \zbf^{(0)} = \betabf^0$, $\ubf^{(0)} = \zerobf$, $\tau>0$ \;
\Repeat{convergence criterion is met}{
$\betabf^{(t+1)} \gets \text{argmin} \, \frac 12 \|\ybf - \Xbf \betabf\|_2^2 + \frac{1}{2\tau} \|\betabf + \zbf^{(t)} + \ubf^{(t)}\|_2^2 + \rho \|\betabf\|_1$\;
$\zbf^{(t+1)} \gets \text{proj}_{\cal C}(\betabf^{(t+1)}+\ubf^{(t)})$\;
$\ubf^{(t+1)} \gets \ubf^{(t)} + \betabf^{(t+1)} + \zbf^{(t+1)}$\;
}
\caption{ADMM for solving the constrained lasso \eqref{eqn:LS-constrlasso}.}
\label{algo:ADMM-constrlasso}
\end{algorithm}

%% Path %%
\subsection{Path Algorithm}
\label{sec:path}

In this section we derive a novel solution path algorithm for the constrained lasso problem \eqref{eqn:LS-constrlasso}.  According to the KKT conditions, the optimal point $\betabf(\rho)$ is characterized by the stationarity condition
\begin{eqnarray*}
	- \Xbf^T [\ybf - \Xbf \betabf(\rho)]  + \rho \sbf(\rho) + \Abf^T \lambdabf(\rho) + \Cbf^T \mubf(\rho) = \zerobf_p
\end{eqnarray*}
coupled with the linear constraints. Here $\sbf(\rho)$ is the subgradient $\partial \|\betabf\|_1$ with elements
\begin{eqnarray}
	s_j(\rho) =  \begin{cases} 1 & \beta_j(\rho)>0 \\ [-1,1] & \beta_j(\rho)=0 \\ -1 & \beta_j(\rho) < 0 \end{cases}, \label{eq:subgrad}
\end{eqnarray}
and $\mubf$ satisfies the complementary slackness condition.  That is, $\mu_l =0$ if $\cbf_l^T \betabf < d_l$ and $\mu_l \ge 0$ if $\cbf_l^T \betabf = d_l$.

Along the path we need to keep track of two sets,
\begin{eqnarray*}
	{\cal A} &:=& \{j: \beta_j \ne 0\}, \quad
	{\cal Z}_I := \{l: \cbf_l^T \betabf = d_l\}.
\end{eqnarray*}
The first set indexes the non-zero (active) coefficients and the second keeps track of the set of (binding) inequality constraints on the boundary.  Focusing on the active coefficients for the time being, we have the (sub)vector equation
\begin{eqnarray}
	\zerobf_{|{\cal A}|} &=& - \Xbf_{:,{\cal A}}^T (\ybf - \Xbf_{:,{\cal A}} \betabf_{\cal A})  + \rho \sbf_{\cal A} + \Abf_{:,{\cal A}}^T \lambdabf + \Cbf_{{\cal Z}_I,{\cal A}}^T \mubf_{{\cal Z}_I} \label{eqn:stationarity-active} \\
	\begin{pmatrix} \bbf \\ \dbf_{{\cal Z}_I} \end{pmatrix} &=& \begin{pmatrix} \Abf_{:,\cal A} \\ \Cbf_{{\cal Z}_I,{\cal A}} \end{pmatrix} \betabf_{\cal A}, \nonumber
\end{eqnarray}
involving dependent unknowns $\betabf_{\cal A}$, $\lambdabf$, and $\mubf_{\Zbf_I}$, and independent variable $\rho$. Applying the implicit function theorem to the vector equation \eqref{eqn:stationarity-active} yields the path following direction
\begin{eqnarray}
	\frac{d}{d \rho} \begin{pmatrix} \betabf_{\cal A} \\ \lambdabf \\ \mubf_{{\cal Z}_I} \end{pmatrix} = - \begin{pmatrix} \Xbf_{:,{\cal A}}^T \Xbf_{:,{\cal A}} & \Abf_{:,{\cal A}}^T & \Cbf_{{\cal Z}_I,{\cal A}}^T \\ \Abf_{:,{\cal A}} & \zerobf & \zerobf \\   \Cbf_{{\cal Z}_I,{\cal A}} & \zerobf & \zerobf \end{pmatrix}^{-1} \begin{pmatrix} \sbf_{\cal A} \\ \zerobf \\ \zerobf \end{pmatrix}. \label{eqn:path-direction}
\end{eqnarray}
The right hand side is constant on a path segment as long as the sets ${\cal A}$ and ${\cal Z}_I$ and the signs of the active coefficients $\sbf_{\cal A}$ remain unchanged. This shows that the solution path of the constrained lasso is piecewise linear. The involved matrix is non-singular as long as $X_{:,{\cal A}}$ has full column rank and the constraint matrix $\begin{pmatrix} \Abf_{:,\cal A} \\ \Cbf_{{\cal Z}_I,{\cal A}} \end{pmatrix}$ has linearly independent rows. The stationarity condition restricted to the inactive coefficients
\begin{eqnarray*}
	- \Xbf_{:,{\cal A}^c}^T [\ybf - \Xbf_{:,{\cal A}} \betabf_{\cal A}(\rho)]  + \rho \sbf_{{\cal A}^c}(\rho) + \Abf_{:,{\cal A}^c}^T \lambdabf(\rho) + \Cbf_{{\cal Z}_I,{\cal A}^c}^T \mubf_{{\cal Z}_I}(\rho) &=& \zerobf_{|{\cal A}^c|}
\end{eqnarray*}
determines
\begin{eqnarray}
	\rho s_{{\cal A}^c}(\rho) = \Xbf_{:,{\cal A}^c}^T [\ybf - \Xbf_{:,{\cal A}} \betabf_{\cal A}(\rho)] - \Abf_{:,{\cal A}^c}^T \lambdabf(\rho) - \Cbf_{{\cal Z}_I,{\cal A}^c}^T \mubf_{{\cal Z}_I}(\rho).	\label{eqn:subgradInact}
\end{eqnarray}
Thus $\rho \sbf_{{\cal A}^c}$ moves linearly along the path via
\begin{eqnarray}
	\frac{d}{d \rho} [\rho \sbf_{{\cal A}^c}] = - \begin{pmatrix} \Xbf_{:,{\cal A}}^T \Xbf_{:,{\cal A}^c}  \\ \Abf_{:,{\cal A}^c} \\ \Cbf_{{\cal Z}_I,{\cal A}^c} \end{pmatrix}^T \frac{d}{d \rho} \begin{pmatrix} \betabf_{\cal A} \\ \lambdabf \\ \mubf_{{\cal Z}_I} \end{pmatrix}.	\label{eqn:path-direction-subgrad}
\end{eqnarray}
The inequality residual $\rbf_{{\cal Z}_I^c} := \Cbf_{{\cal Z}_I^c,{\cal A}} \betabf_{\cal A} - \dbf_{{\cal Z}_I^c}$ also moves linearly with gradient
\begin{eqnarray}
	\frac{d}{d \rho} \rbf_{{\cal Z}_I^c} = \Cbf_{{\cal Z}_I^c,{\cal A}}  \frac{d}{d \rho} \betabf_{\cal A}.	\label{eqn:path-direction-ineqresid}
\end{eqnarray}
Together, equations \eqref{eqn:path-direction}, \eqref{eqn:path-direction-subgrad}, and \eqref{eqn:path-direction-ineqresid} are used to monitor changes to $\cal A$ and ${\cal Z}_I$, which can potentially result in kinks in the solution path.  

To recap, since the solution path is piecewise linear we only need to monitor the events discussed above that can result in kinks along the path, and then the rest of the path can be interpolated.  A summary of these events is given in the column on the left in Table~\ref{tab:events}.  We perform path following in the decreasing direction from $\rho_{\text{max}}$ towards $\rho = 0$.  Let $\betabf^{(t)}$ denote the solution at kink $t$, then the next kink $t + 1$ is identified by the smallest $\Delta \rho$, where $\Delta \rho>0$ is determined by the conditions listed in the right column of Table~\ref{tab:events}.  

\begin{table}[h!]
\caption{Solution Path Events}
\begin{tabular}{cc}
\toprule
\toprule
Event & Monitor \\
\midrule
\small An active coefficient hits 0 & \small $\betabf_{\cal A}^{(t)} - \Delta \rho \frac{d}{d \rho} \betabf_{\cal A}^{(t)} = \zerobf_{|{\cal A}|}$ \\
\small An inactive coefficient becomes active & \small $[\rho^{(t)} \sbf_{{\cal A}^c}^{(t)}] - \Delta \rho \frac{d}{d \rho} [\rho \sbf_{{\cal A}^c}] = \pm (\rho^{(t)} - \Delta \rho) \onebf_{|{\cal A}^c|}$ \\
\small A strict inequality constraint hits the boundary & \small $\rbf_{{\cal Z}_I^c}^{(t)} - \Delta \rho \frac{d}{d \rho} \rbf_{{\cal Z}_I^c} = \zerobf_{|{\cal Z}_I^c|}$	\\
\small An inequality constraint escapes the boundary & \small $\mubf_{{\cal Z}_I}^{(t)} - \Delta \rho \frac{d}{d \rho} \mubf_{{\cal Z}_I} = \zerobf_{|{\cal Z}_I|}$	\\ 
\bottomrule
\end{tabular}
\label{tab:events}
\end{table}

In addition to monitoring these events along the path, we also need to ensure that the subgradient conditions \eqref{eq:subgrad} remain satisfied.  An issue arises when inactive coefficients on the boundary of the subgradient interval are moving too slowly along the path such that their subgradient would escape [-1, 1] at the next kink $t + 1$.  To handle this issue, if an inactive coefficient $\beta_j$, $j \in {\cal A}^c$, with subgradient $s_j = \pm 1$ is moving too slowly, the coefficient is moved to the active set ${\cal A}$ and equation \eqref{eqn:path-direction} is recalculated before continuing the path algorithm.  See Appendix~\ref{sec:subgrad} for the explicit ranges of $\frac{d}{d \rho} [\rho \sbf_{{\cal A}^c}]$ that need to be monitored and the corresponding derivations.

\subsubsection{Initialization}

Since we perform path following in the decreasing direction, a starting value for the tuning parameter, $\rho_{\text{max}}$, is needed.  As $\rho \rightarrow \infty$, the solution to the original problem \eqref{eqn:LS-constrlasso} is given by
\begin{eqnarray}
	&\text{minimize}& \quad \|\betabf\|_1 \label{eqn:LP} \\
	&\text{subject to}& \quad \Abf \betabf = \bbf \text{ and } \Cbf \betabf \le \dbf, \nonumber
\end{eqnarray}
which is a standard linear programming problem.  We first solve \eqref{eqn:LP} to obtain initial coefficient estimates $\betabf^0$ and the corresponding sets ${\cal A}$ and ${\cal Z}_I$, as well as initial values for the Lagrange multipliers $\lambdabf^0$ and $\mubf^0$.  Following \citet{rosset2007piecewise}, path following begins from 
\begin{eqnarray}
	\rho_{\text{max}} = \max \,\, \left| \xbf_j^T (\ybf - \Xbf \betabf^0) - \Abf_{:,j}^T\lambdabf^0 - \Cbf_{{\cal Z}_I,j}^T \mubf_{{\cal Z}_I}^0 \right|, \label{eq:rhoMax}
\end{eqnarray}
and the subgradient is set according to \eqref{eq:subgrad} and \eqref{eqn:subgradInact}.  As noted by \citet{JamesPaulsonRusmevichientong14CLasso}, this approach can fail when \eqref{eqn:LP} does not have a unique solution.  For example, consider a constrained lasso with a sum-to-one constraint on the coefficients, $\sum_j \beta_j = 1$.  Any elementary vector $\ebf_j$, which has a 1 for the $j^{th}$ element and 0 otherwise, satisfies this constraint while also achieving the minimum $\ell_1$ norm, resulting in multiple solutions to \eqref{eqn:LP}.  In this case, it is still possible to use \eqref{eqn:LP} and \eqref{eq:rhoMax} to identify $\rho_{\text{max}}$, which is then used in \eqref{eqn:LS-constrlasso} to initialize $\betabf^0$, ${\cal A}$, ${\cal Z}_I$, $\lambdabf^0$, and $\mubf^0$ via quadratic programming.

\section{Degrees of Freedom}
\label{sec:df}
Now that we have studied several methods for solving the constrained lasso, we turn our attention to deriving a formula for degrees of freedom.  The standard approach in the lasso literature \citep{efron2004lars, zou2007degrees, TibshiraniTaylor11GenLasso, tibshirani2012degrees} is to rely on the expression for degrees of freedom given by \citet{stein81},
\begin{eqnarray}
	\text{df}(g) = \E \left[  \sum_{i = 1}^n  \frac{\partial g_i}{\partial y_i}  \right] , \label{eq:df}
\end{eqnarray}
where $g$ is a continuous and almost differentiable function, which with $g(y) = \hat{\ybf} = \Xbf \hat \betabf$ is satisfied in our case \citep{Hu2015}.  In order to apply \eqref{eq:df}, we need to assume that the response is normally distributed, 
$$
\ybf \sim N(\mubf, \sigma^2 \Ibf).  
$$
As before, we also assume that both constraint matrices, $\Abf$ and $\Cbf$, have full row rank, and $\Xbf$ has full column rank.  Then, using the results in \citet{Hu2015} with $\Dbf = \Ibf$, for a fixed $\rho \ge 0$ the degrees of freedom are given by
\begin{eqnarray}
	\text{df}(\Xbf \hat{\betabf}(\rho)) = \E \left[  | \mathcal{A} | - (q + | \mathcal{Z}_I |) \right] , \label{eq:classoDF}
\end{eqnarray}
where $| \mathcal{A} |$ is the number of active predictors, $q$ is the number of equality constraints, and $| \mathcal{Z}_I |$ is the number of binding inequality constraints.  The unbiased estimator for the degrees of freedom is then $ | \mathcal{A} | - (q + | \mathcal{Z}_I |)$.  This result is intuitive as the degrees of freedom starts out as the number of active predictors, and then one degree of freedom is lost for each equality constraint and each binding inequality constraint.  Additionally, when there are no constraints, \eqref{eqn:LS-constrlasso} becomes a standard lasso problem with degrees of freedom equal to $| \mathcal{A} |$, consistent with the result in \citet{zou2007degrees}.  The formula \eqref{eq:classoDF} is also consistent with results for constrained estimation presented in \citet{ZhouLange2013Constrained} and \citet{ZhouWu13EPSODE}.  Degrees of freedom are an important measure that is an input for several metrics used for model assessment and selection, such as Mallows' $C_p$, AIC, and BIC.  Specifically, these criteria can be plotted along the path as a function of $\rho$ as a technique for selecting the optimal value for the tuning parameter, as alternatives to cross-validation.  Degrees of freedom are also used in implementing the solution path algorithm, as the path terminates once the degrees of freedom equal $n$.

\section{Simulated Examples}
\label{sec:sims}
To investigate the performance of the various algorithms outlined in Section \ref{sec:algo} for solving a constrained lasso problem, we consider two simulated examples.  For both simulations, we used the three different algorithms discussed in Section~\ref{sec:algo} to solve \eqref{eqn:LS-constrlasso}.  As noted in Section~\ref{sec:ADMM}, the ADMM algorithm includes an additional tuning parameter $\tau$, which we fix at $1/n$ based on initial experiments.  Additionally, as pointed out in \citet{Boyd11ADMM}, the performance of the ADMM method can be greatly impacted by the choice of the algorithm's stopping criteria, which we set to be $10^{-4}$ for both the absolute and relative error tolerances.  We also use a user-defined function handle to solve the subproblem of projecting onto the constraint set for ADMM as this improves efficiency.  Two factors of interest in the simulations are the size of the problem, $(n,p)$, and the value of the regularization tuning parameter, $\rho$.  Four different levels were used for the size factor, $(n,p)$: (50, 100), (100, 500), (500, 1000), and (1000, 2000).  For the latter factor, the values of $\rho$ were calculated as a fraction of the maximum $\rho$.  The fractions, or $\rho_{\text{scale}}$ values,\footnote{I.e., $\rho = \rho_{\text{scale}} \cdot \rho_{\text{max}}$.} used in the simulations were 0.2, 0.4, 0.6, and 0.8, to investigate how the degree of regularization impacts algorithm performance.  Since the runtimes for quadratic programming and ADMM are for a fixed value of $\rho$, the total runtime for the solution path algorithm is averaged across the number of kinks in the path to make the results more comparable.  To generate the data for both simulations, the covariates in the design matrix, $\Xbf$, were generated as independent and identical (iid) standard normal variables, and the response was generated as  $\ybf = \Xbf \betabf + \varepsilonbf$ where $\varepsilonbf \sim N(\zerobf, \Ibf)$.  Both simulations used 20 replicates and were conducted in \textsc{Matlab} using the {\tt SparseReg} toolbox on a computer with an Intel i7-6700 3.4 GHz processor and 32 GB memory.  Quadratic programming uses the {\sc Gurobi Optimizer} via the \textsc{Matlab} interface, while ADMM and the path algorithm are pure \textsc{Matlab} implementations. 

\subsection{Sum-to-zero Constraints}
 The first simulation involves a sum-to-zero constraint on the true parameter vector, $\sum_j \beta_j = 0$.  Recently, this type of constraint on the lasso has seen increased interest as it has been used in the analysis of compositional data as well as analyses involving any biological measurement analyzed relative to a reference point \citep{lin2014varselectcompositional, shi2016microbiome, Altenbuchinger2016}.  Written in the constrained lasso formulation \eqref{eqn:LS-constrlasso}, this corresponds to $\Abf = \onebf_p^T$ and $\bbf = 0$.  For this simulation, the true parameter vector, $\betabf$, was defined such that the first 25\% of the entries are 1, the next 25\% of the entries are -1, and the rest of the elements are 0.  Thus the true parameter satisfies the sum-to-zero constraint, which we can impose on the estimation using the constraints.  

\begin{figure}[h!]
	\begin{center}
		\includegraphics[scale=0.8]{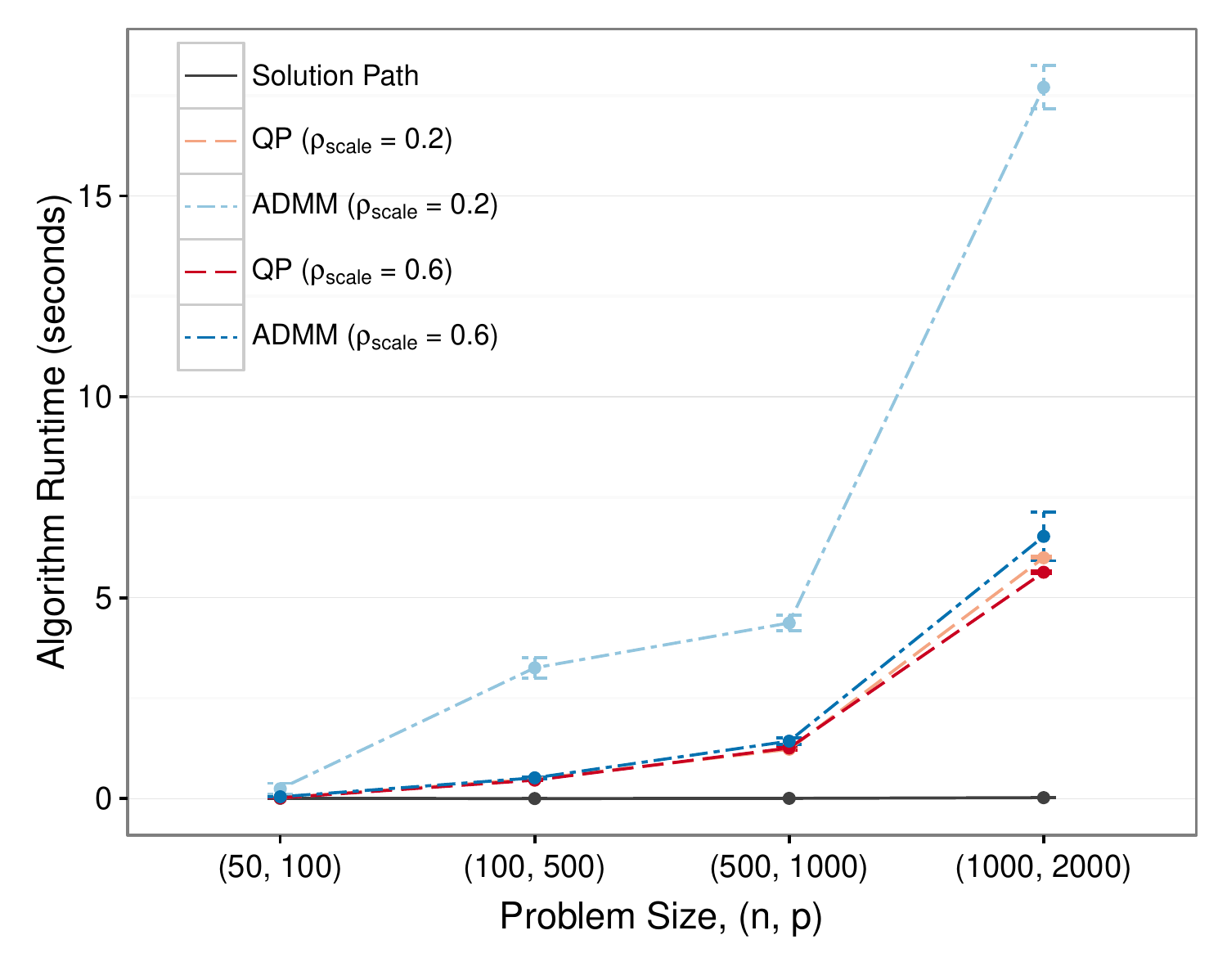} 
	\caption{Simulation 1 Results. \small Average algorithm runtime (seconds) plus/minus one standard error for a constrained lasso with sum-to-zero constraints on the coefficients.  The solution path's runtime is averaged across the number of kinks in the path to make the runtime more comparable to the other algorithms estimated at one value of the tuning parameter, $\rho = \rho_{\text{scale}} \cdot \rho_{\text{max}}$.}
	\label{fig:sim1Time}
	\end{center}
\end{figure}

The main results of the simulation are given in Figure~\ref{fig:sim1Time}, which plots the average algorithm runtime results across different problem sizes, $(n, p)$.  The results using quadratic programing (QP) and ADMM are each graphed at two values of $\rho_{\text{scale}}$, 0.2 and 0.6.  In the graph we can see that the solution path algorithm performed faster than the other methods, and its relative performance is even more impressive as the problem size grows.  The graph also shows the impact of the tuning parameter, $\rho$, on both QP and ADMM.  QP performed similarly across both values of $\rho_{\text{scale}}$, but that was not the case for ADMM.  At $\rho_{\text{scale}} = 0.6$, ADMM performed  very similarly to QP, but ADMM's performance was much worse at smaller values of $\rho$.  Smaller values of $\rho$ correspond to less weight on the $\ell_1$ penalty, which results in solutions that are less sparse.  The ADMM runtime is also more variable than the other algorithms.  

While algorithm runtime is the metric of primary interest, a fast algorithm is not of much use if it is woefully inaccurate.  Figure~\ref{fig:sim1error} in Appendix~\ref{sec:appFigs} plots the objective value error relative to QP for the solution path and ADMM for $(n, p) = (500, 1000)$.  The results from the other problem sizes are qualitatively similar and are thus omitted.  From Figure~\ref{fig:sim1error}, we see that the solution path algorithm is not only efficient but also accurate.  On the other hand, the accuracy of ADMM decreases as $\rho$ increases.  Part of this is to be expected given that convergence tolerance used for ADMM is less stringent than the one used for QP.  However, the magnitude of these errors is probably not of practical importance.

\subsection{Non-negativity Constraints}
The second simulation involves the positive lasso mentioned in Section~\ref{sec:intro}, as to our knowledge it is the most common version of the constrained lasso that has appeared in the literature.  Also referred to as the non-negative lasso, as its name suggests it constrains the lasso coefficient estimates to be non-negative.  In the constrained lasso formulation \eqref{eqn:LS-constrlasso}, the positive lasso corresponds to the constraints $\Cbf = -\Ibf_p$ and $\dbf = \zerobf_p$.  For each problem size, the true parameter vector was defined as
\begin{eqnarray*}
	\beta_j =  \begin{cases} j, & j = 1, ..., 10 \\ 0, & j = 11, ..., p \end{cases}, 
\end{eqnarray*}
so the true coefficients obey the constraints and the constrained lasso allows us to incorporate this prior knowledge into the estimation.   

 \begin{figure}[h!]
	\begin{center}
		\includegraphics[scale=0.8]{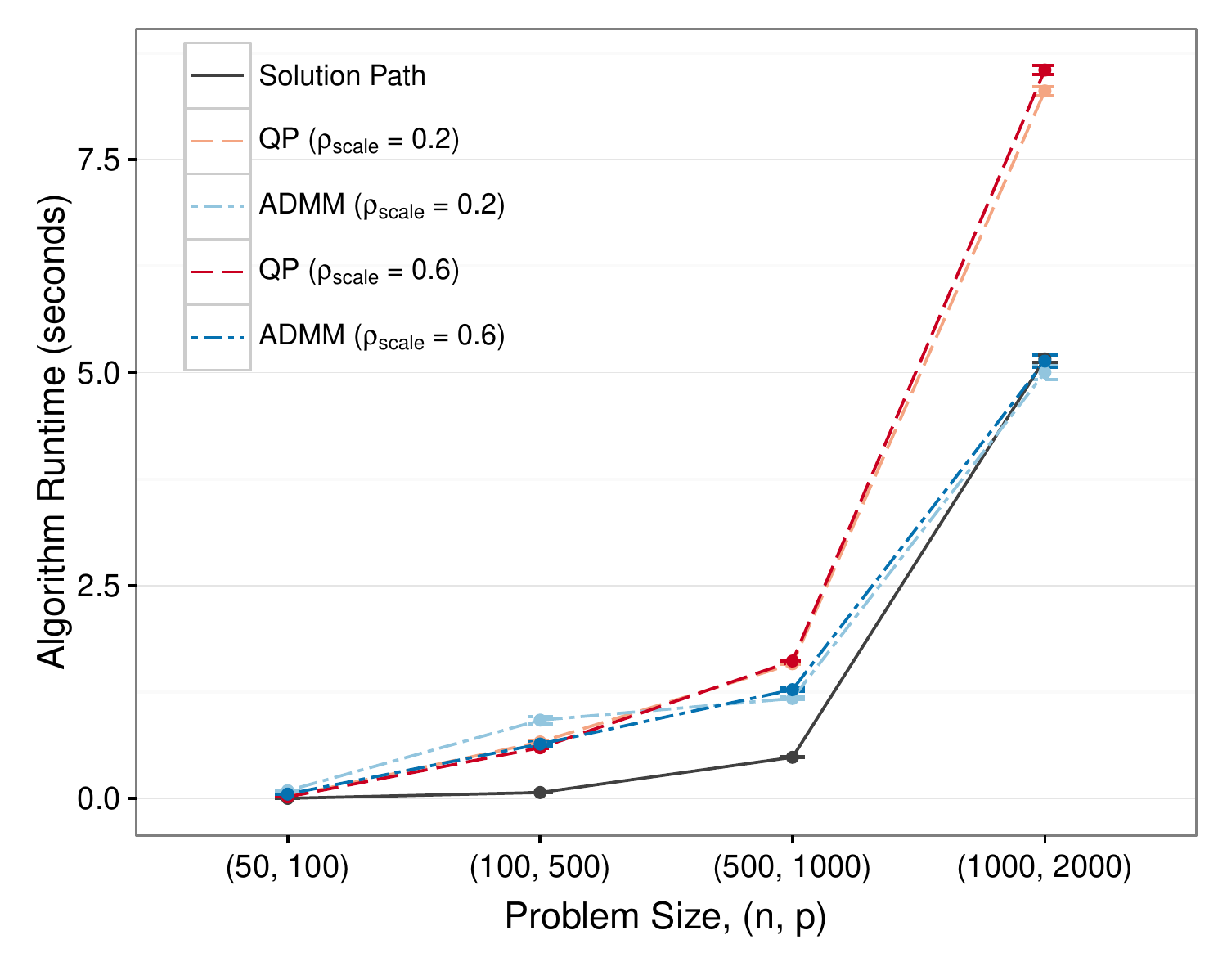} 
	\caption{Simulation 2 Results. \small Average algorithm runtime (seconds) plus/minus one standard error for a constrained lasso with non-negativity constraints on the parameters.  The solution path's runtime is averaged across the number of kinks in the path to make the runtime more comparable to the other algorithms which are estimated at one value of the tuning parameter, $\rho = \rho_{\text{scale}} \cdot \rho_{\text{max}}$. }
	\label{fig:sim2Time}
	\end{center}
\end{figure}

Figure~\ref{fig:sim2Time} is a graph of the average runtime for each algorithm for the different problem sizes considered.  As with simulation 1, the results for quadratic programming (QP) and ADMM are graphed at two different values of $\rho$, corresponding to $\rho_{\text{scale}} =  \frac{\rho}{\rho_{\text{max}}} \in \{ 0.2, 0.6\}$ to also demonstrate the impact of $\rho$ on the estimation time.  One noteworthy result is that ADMM fared better relative to QP as the problem size grew and was faster than QP for the two larger sizes under investigation, whereas ADMM had runtimes that were comparable or slower than QP in the first simulation.  We expected ADMM to outperform QP for larger problems, which happened more quickly in this setting since the inclusion of $p$ inequality constraints notably increased the complexity of the problem.  As with the first simulation, another thing that stands out in the results is the strong performance of the solution path algorithm, which generally outperformed the other two methods.  However, for $(n, p) = (1000, 2000)$, ADMM and the solution path performed similarly, partly due to the initialization of the path algorithm hampering its performance as the problem size grows.  In terms of accuracy, the objective value errors relative to QP for the solution path and ADMM were negligible and are thus omitted.  

\section{Real Data Applications}
\label{sec:realData}
\subsection{Global Warming Data}
\begin{figure}[h!]
	\begin{center}
		\includegraphics[scale=0.8]{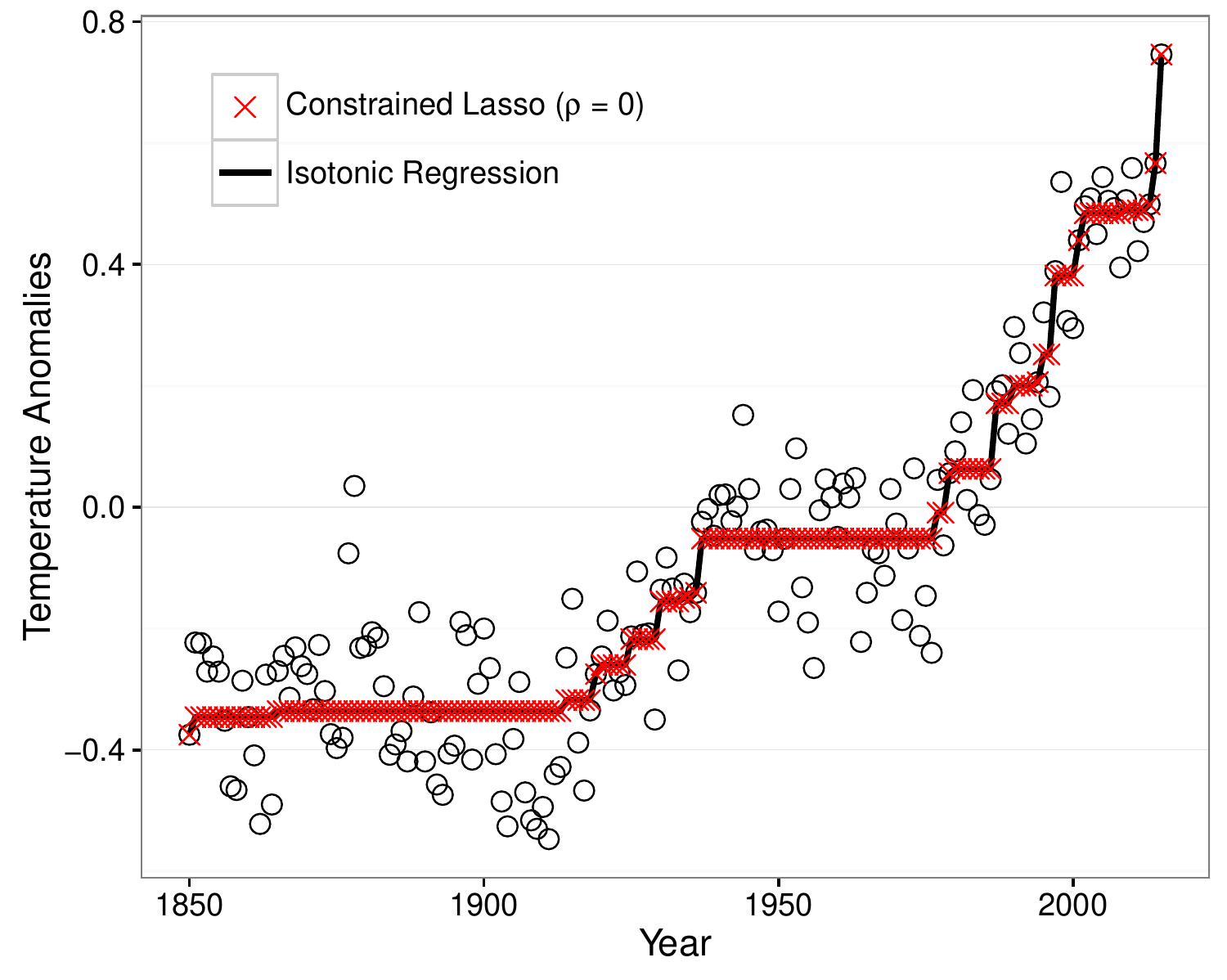} % exported as 6x4.8'' pdf from R
	\caption{Global Warming Data. \small Annual temperature anomalies relative to the 1961-1990 average, with trend estimates using isotonic regression and the constrained lasso.}
	\label{fig:warmingEstimates}
	\end{center}
\end{figure}

For our first application of the constrained lasso on a real dataset, we revisit the global temperature data presented in Section~\ref{sec:intro}, which was provided by \citet{jones2016global}.\footnote{The dataset can be downloaded from the \href{http://cdiac.ornl.gov/trends/temp/jonescru/jones.html}{Carbon Dioxide Information Analysis Center} at the Oak Ridge National Laboratory.}  The dataset consists of annual temperature anomalies from 1850 to 2015, relative to the average for 1961-90.  As mentioned, there appears to be a monotone trend to the data over time, so it is natural to want to incorporate this information when the trend is estimated.  \citet{wu2001isotonic} achieved this on a previous version of the dataset by using isotonic regression, which is given by  
\begin{eqnarray}
	&\text{minimize}& \quad \frac 12 \| \ybf - \betabf\|_2^2 \label{eq:isotone} \\
	&\text{subject to}& \quad \beta_1 \le \cdots \le \beta_n,  \nonumber
\end{eqnarray}
where $\ybf \in \reals^n$ is the monotonic data series of interest and $\betabf \in \reals^n$ is a monotonic sequence of coefficients.  The lasso analog of isotonic regression,\footnote{Note that the \textit{monotone lasso} discussed by \citet{hastie2007forward} refers to a monotonicity of the solution paths for each coefficient across $\rho$, not a monotonic ordering of the coefficients at each value of the tuning parameter. } which adds an $\ell_1$ penalty term to \eqref{eq:isotone}, can be estimated by the constrained lasso \eqref{eqn:LS-constrlasso} using the constraint matrix 

\begin{eqnarray*}
	\Cbf = \begin{pmatrix} 
	1 & -1 \\
	& 1 & -1 \\
	& & \ddots & \ddots \\
	& & & & 1 & -1 
	\end{pmatrix}
\end{eqnarray*}
and $\dbf = \zerobf$.  In this formulation, isotonic regression is a special case of the constrained lasso with $\rho = 0$.  This is verified by Figure~\ref{fig:warmingEstimates}, which plots the constrained lasso fit at $\rho = 0$ with the estimates using isotonic regression.  

\subsection{Brain Tumor Data}

\begin{figure}[h!]
	\begin{center}
		\includegraphics[scale=0.8]{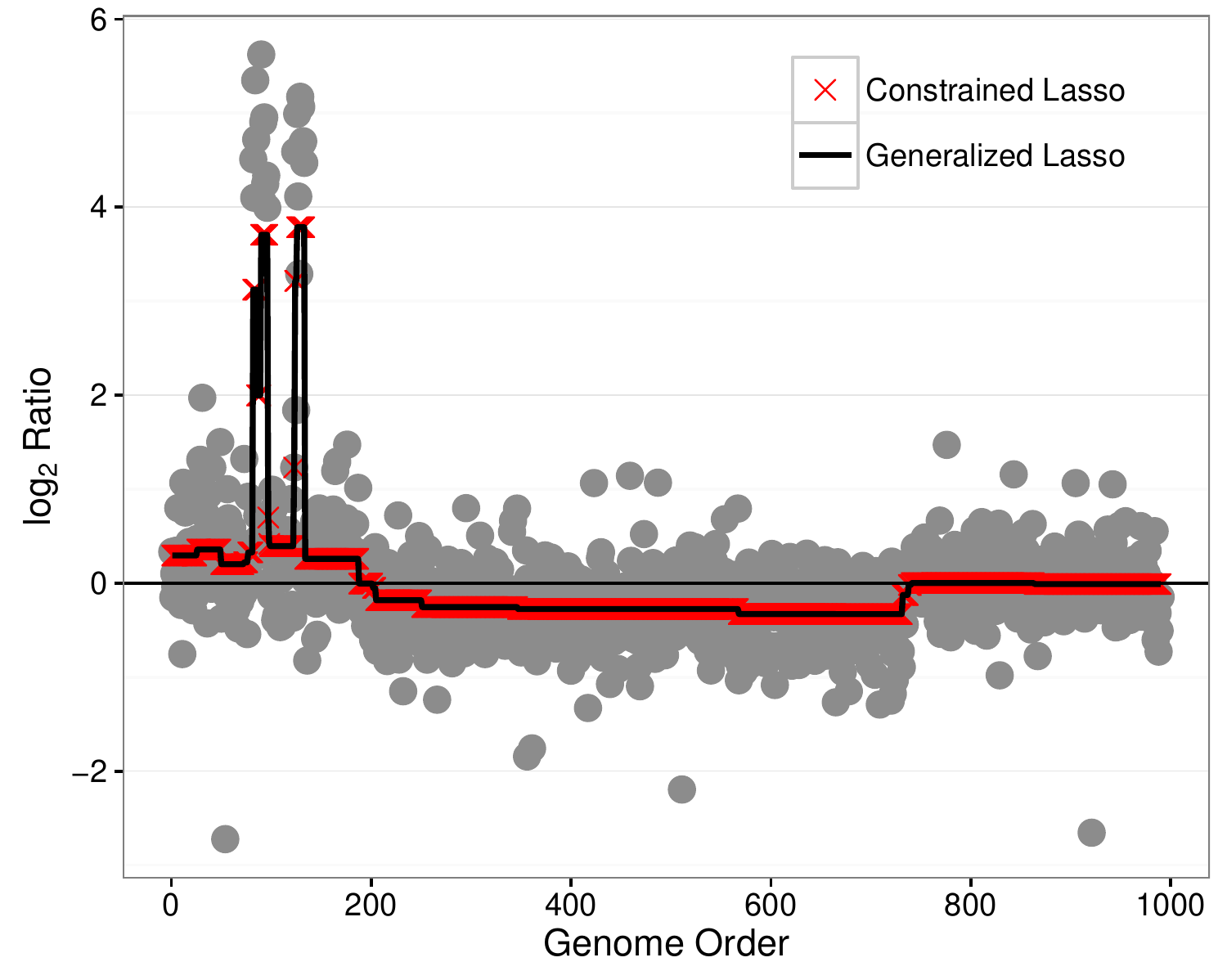} % exported as 6x4.8'' pdf from R
	\caption{Brain Tumor Data. \small Sparse fused lasso estimates on the brain tumor data using both the generalized lasso and the constrained lasso.}
	\label{fig:cghEstimates}
	\end{center}
\end{figure}

Our second application of the constrained lasso uses a version of the comparative genomic hybridization (CGH) data from \citet{bredel2005gbm} that was modified and studied by \citet{tibshirani2008cgh},\footnote{This version of the dataset is available in the \href{https://cran.r-project.org/web/packages/cghFLasso/index.html}{{\tt cghFLasso} R package.}} which can be seen in Figure~\ref{fig:cghEstimates}.  The dataset contains CGH measurements from 2 glioblastoma multiforme (GBM) brain tumors.  CGH array experiments are used to estimate each gene's DNA copy number by obtaining the $\log_2$ ratio of the number of DNA copies of the gene in the tumor cells relative to the number of DNA copies in the reference cells.  Mutations to cancerous cells result in amplifications or deletions of a gene from the chromosome, so the goal of the analysis is to identify these gains or losses in the DNA copies of that gene \citep{michels2007detection}.  \citet{tibshirani2008cgh} proposed using the sparse fused lasso to approximate the CGH signal by a sparse, piecewise constant function in order to determine the areas with non-zero values, as positive (negative) CGH values correspond to possible gains (losses).  The sparse fused lasso \citep{tibshirani2005fused} is given by 
\begin{eqnarray}
	&\text{minimize}& \quad \frac 12 \| \ybf - \betabf\|_2^2 + \rho_1 \| \betabf\|_1 + \rho_2 \sum_{j=2}^p | \beta_j - \beta_{j-1}|, \label{eqn:sflasso}
\end{eqnarray}
where the additional penalty term encourages the estimates of neighboring coefficient to be similar, resulting in a piecewise constant function.  This modification of the lasso was originally termed the fused lasso, but in line with \citet{TibshiraniTaylor11GenLasso} we refer to \eqref{eqn:sflasso} as the sparse fused lasso to distinguish it from the related problem that does not include the sparsity-inducing $\ell_1$ norm on the coefficients.  Regardless, the sparse fused lasso is a special case of the generalized lasso \eqref{eqn:LS-genlasso} with the penalty matrix
\begin{eqnarray*}
	\Dbf = \begin{pmatrix} 
	-1 & 1 \\
	& -1 & 1 \\
	& & \ddots & \ddots \\
	& & & & -1 & 1 \\
	1 & \\
	& 1 \\
	& & \ddots \\
	& & & & 1 \\
	& & & & & 1
	\end{pmatrix} \in \mathbb{R}^{(2p-1) \times p}.
\end{eqnarray*}
As discussed in Section~\ref{sec:genlasso2classo}, the sparse fused lasso can be reformulated and solved as a constrained lasso problem.  Estimates of the underlying CGH signal from solving the sparse fused lasso as both a generalized lasso (using the \href{https://github.com/statsmaths/genlasso}{{\tt genlasso} R package} \citep{Tibshirani14genlassoR}) and a constrained lasso are given in Figure~\ref{fig:cghEstimates}.  As can be seen, the estimates from the two different methods match, providing empirical verification of the transformation derived in Section~\ref{sec:genlasso2classo}.  

\subsection{Microbiome Data}
Our last real data application with the constrained lasso uses microbiome data.  The analysis of the human microbiome, which consists of the genes from all of the microorganisms in the human body, has been made possible by the emergence of next-generation sequencing technologies.  Microbiome research has garnered much interest as these cells play an important role in human health, including energy levels and diseases; see \citet{li2015microbiome} and the references therein.  Since the number of sequencing reads varies greatly from sample to sample, often the counts are normalized to represent the relative abundance of each bacterium, resulting in compositional data, which are proportions that sum to one.  Motivated by this, regression \citep{shi2016microbiome} and variable selection \citep{lin2014varselectcompositional} tools for compositional covariates have been developed, which amount to imposing sum-to-zero constraints on the lasso.  

\citet{Altenbuchinger2016} built on this work by demonstrating that a sum-to-zero constraint is useful anytime the normalization of data relative to some reference point results in proportional data, as is often the case in biological applications, since the analysis using the constraint is insensitive to the choice of the reference.  \citet{Altenbuchinger2016} derived a coordinate descent algorithm for the elastic net with a zero-sum constraint, 
\begin{eqnarray}
	&\text{minimize}& \quad \frac 12 \| \ybf - \Xbf \betabf\|_2^2 + \rho \left(\alpha \|\betabf\|_1 + \frac{1 - \alpha}{2} \| \betabf \|_2^2 \right) \label{eqn:conEnet} \\
	&\text{subject to}& \quad \sum_j \beta_j = 0,  \nonumber
\end{eqnarray}
but the focus of their analysis, which they refer to as \textit{zero-sum regression}, corresponds to $\alpha = 1$, for which \eqref{eqn:conEnet} reduces to the constrained lasso \eqref{eqn:LS-constrlasso}.  \citet{Altenbuchinger2016} applied zero-sum regression to a microbiome dataset from \citet{weber2015low} to demonstrate zero-sum regression's insensitivity to the reference point, which was not the case for the regular lasso.  The data contains the microbiome composition of patients undergoing allogeneic stem cell transplants (ASCT) as well as their urinary levels of 3-indoxyl sulfate (3-IS), a metabolite of the organic compound indole that is produced in the colon and liver.  ASCT patients are at high risk for acute graft-versus-host disease and other infectious complications, which have been associated with the composition of the microbiome and the absence of protective microbiota-born metabolites in the gut \citep{taur2012intestinal, holler2014metagenomic, murphy2011role}.  One such protective substance is indole, which is a byproduct when gut bacteria breaks down the amino acid tryptophan \citep{weber2015low}.

\begin{figure}%
    \centering
    \subfloat{{\includegraphics[scale=0.55]{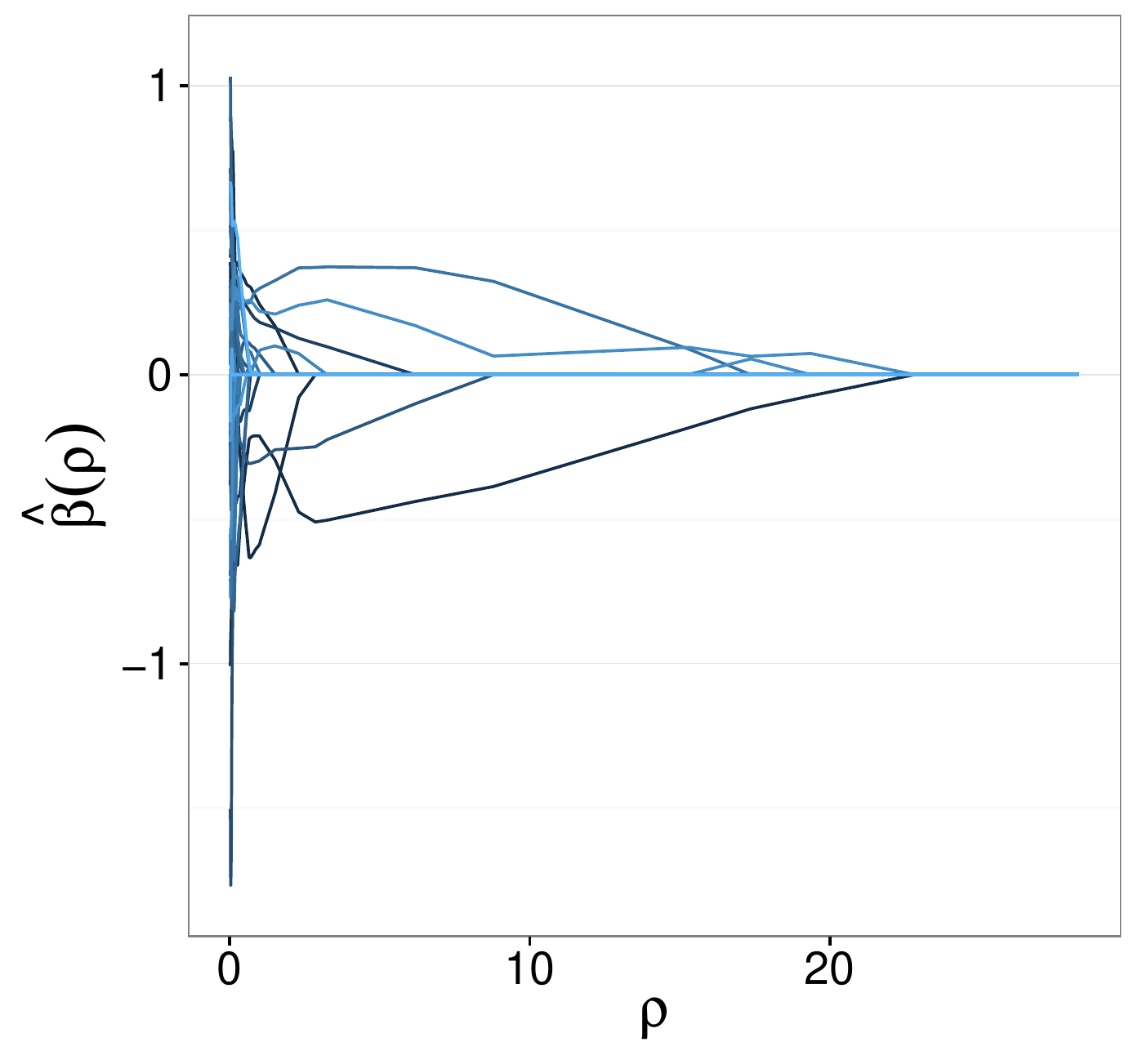} }}%
    \subfloat{{\includegraphics[scale=0.55]{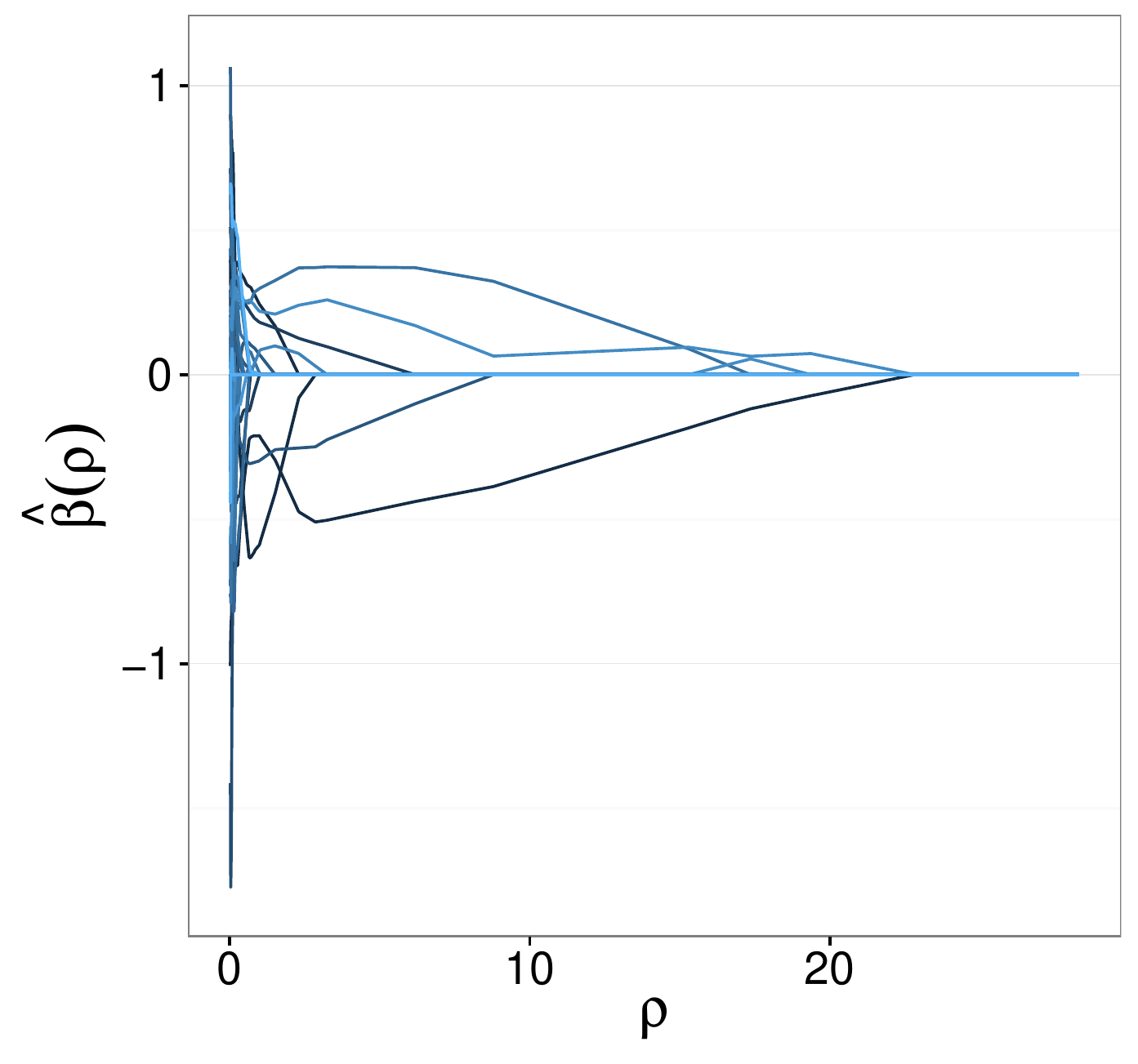} }}%
    \caption{Microbiome Data Solution Paths. \small Comparison of solution path coefficient estimates on the microbiome dataset using both zero-sum regression (left panel) and the constrained lasso (right panel). }%
    \label{fig:microbiome}%
\end{figure}

Of interest, then, is to identify a small subset of the microbiome composition associated with 3-IS levels, as the presence of relatively more indole-producing bacteria in the intestines is expected to result in higher levels of 3-IS in urine.  ASCT patients receive antibiotics that kill gut bacteria, but with a better understanding of which bacteria produce indole, antibiotics that spare those bacteria could be used instead \citep{Altenbuchinger2016}.  The dataset itself contains information on 160 bacteria genera from 37 patients.\footnote{The dataset, as well as the code to reproduce the results in \citet{Altenbuchinger2016}, are available in the \href{https://github.com/rehbergT/zeroSum}{{\tt zeroSum} R package} on Github.}  The bacteria counts were $\log_2$-transformed and normalized to have a constant average across samples.   Figure~\ref{fig:microbiome} plots $\hat{\betabf}(\rho)$, the coefficient estimate solution paths as a function of $\rho$, using both zero-sum regression and the constrained lasso.  As can be seen in the graphs, the coefficient estimates are nearly indistinguishable except for some very minor differences, which are a result of the slightly different formulations of the two problems.  Since this is a case where $n < p$, a small ridge penalty is added to the constrained lasso objective function \eqref{eqn:LS-constrlassoRidge} as discussed in Section~\ref{sec:algo}, but unlike \eqref{eqn:conEnet}, the weight on the $\ell_2$ penalty does not vary across $\rho$.

\section{Conclusion}
\label{sec:conclusion}

We have studied the constrained lasso problem, in which the original lasso problem is expanded to include linear equality and inequality constraints.  As we have discussed and demonstrated through real data applications, as well as other examples cited from the literature, the constraints allow users to impose prior knowledge on the coefficient estimates.  Additionally, we have shown that another flexible lasso variant, the generalized lasso, can always be reformulated and solved as a constrained lasso, which greatly enlarges the trove of problems the constrained lasso can solve.  

We derived and compared three different algorithms for computing the constrained lasso solutions as a function of the tuning parameter $\rho$:  quadratic programing (QP), the alternating direction method of multipliers (ADMM), and a novel derivation of a solution path algorithm.  Through simulated examples, it was shown that the solution path algorithm outperforms the other methods in terms of estimation time, without sacrificing accuracy.  For modest problem sizes, QP remained competitive with ADMM, but ADMM became relatively more efficient as the problem size increased.  However, the performance of ADMM is more sensitive to the particular value of $\rho$.  \textsc{Matlab} code to implement these algorithms is available in the \href{https://github.com/Hua-Zhou/SparseReg}{\tt SparseReg} toolbox on Github.  

There are several possible extensions that have been left for future research.  The efficiency of the solution path algorithm may be able to be improved by using the sweep operator \citep{goodnight79sweep} to update \eqref{eqn:path-direction} along the path, as was done in related work by \citet{ZhouLange2013Constrained}.  It also may be of interest to extend the algorithms presented here to more general formulations of the constrained lasso.  All of the algorithms can be extended to handle general convex loss functions, such as a negative log likelihood function for a generalized linear model extension, which was already studied by \cite{JamesPaulsonRusmevichientong14CLasso} using a modified coordinate descent algorithm.  In this case, an extension of the solution path algorithm could be tracked by solving a system of ordinary differential equations (ODE) as in \citet{ZhouWu13EPSODE}.    

\section*{Acknowledgments}

The authors thank Colleen McKendry for her helpful comments.  This research is partially supported by National Science Foundation grants DMS-1055210 and DMS-1127914, and National Institutes of Health grants R01 HG006139, R01 GM53275, and R01 GM105785.

\newpage

\begin{appendix}
\section{Appendix}
\subsection{Additional Figures}
\label{sec:appFigs}
\begin{figure}[h!]
	\begin{center}
		\includegraphics[scale=0.8]{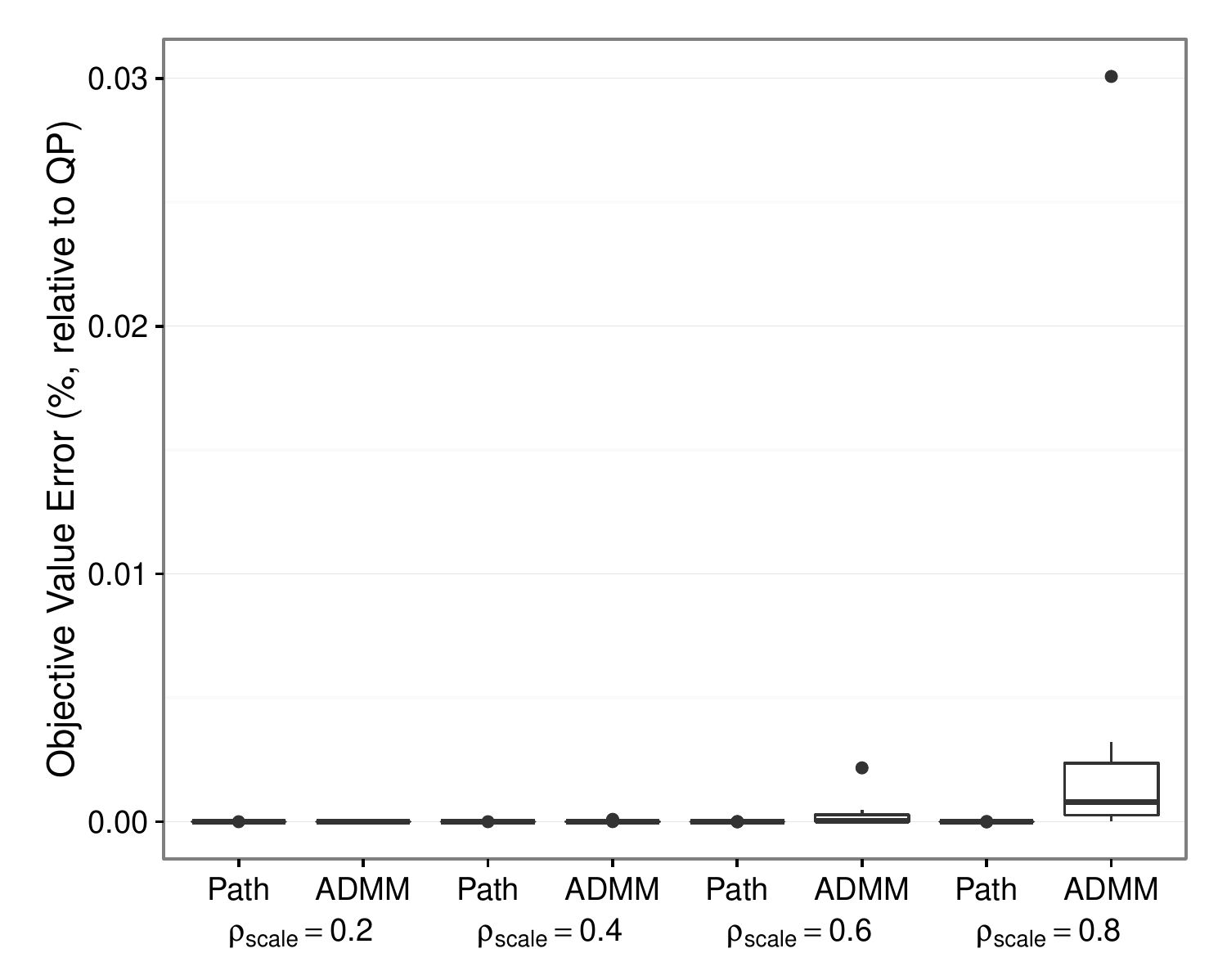} 
	\caption{Objective value error (percent) relative to quadratic programing (QP) for the solution path and ADMM at different values of $\rho_{\text{scale}} =  \frac{\rho}{\rho_{\text{max}}}$ for $(n, p) = (500, 1000)$.  The results are qualitatively the same for the other combinations of $(n, p)$ considered.  }
	\label{fig:sim1error}
	\end{center}
\end{figure}
\subsection{Connection to the Generalized Lasso}
\label{sec:classo2genlasso}
As detailed in Section~\ref{sec:genlasso2classo}, it is always possible to reformulate and solve a generalized lasso as a constrained lasso.  In this section, we demonstrate that it is not always possible to transform a constrained lasso to a generalized lasso.  

\subsubsection{Reparameterization}
However, we first examine a situation where it is in fact possible to transform a constrained lasso to a generalized lasso.  Consider a constrained lasso with only equality constraints and $\bbf = \zerobf$,

\begin{eqnarray}
	&\text{minimize}& \quad \frac 12 \| \ybf - \Xbf \betabf\|_2^2 + \rho \|\betabf\|_1 \label{eqn:LS-constrlassoEqb0} \\
	&\text{subject to}& \quad \Abf \betabf = \zerobf,	\nonumber
\end{eqnarray}
where $\Abf \in \reals^{q \times p}$ with $\rank(\Abf) = q$.  Consider a matrix $\Dbf \in \reals^{p \times p-q}$ whose columns span the null space of $\Abf$. For example, we can use $\Qbf_2$ from the QR decomposition of $\Abf^T$.  Then we can use the change of variables 
\begin{eqnarray}
	 \betabf = \Dbf \thetabf \nonumber,
 \end{eqnarray}
so the objective function becomes 
\begin{eqnarray}
	&\text{minimize}& \quad \frac 12 \| \ybf - \Xbf \Dbf \thetabf\|_2^2 + \rho \|\Dbf \thetabf \|_1 \label{eqn:LS-constrlassoGenlasso},
\end{eqnarray}
and the constraints can be written as 
\begin{eqnarray}
	\Abf \betabf = \Abf \Dbf \thetabf = \zerobf \thetabf  = \zerobf \nonumber.
 \end{eqnarray}
Thus the constraints vanish, as they hold for all $\thetabf$, and we are left with an unconstrained generalized lasso \eqref{eqn:LS-constrlassoGenlasso}.  This result is not surprising in light of the result in Section~\ref{sec:genlasso2classo}, which showed that a generalized lasso reformulated as a constrained lasso has the constraints $\Ubf_2^T \alphabf = \zerobf_{m-r}$.  That is a case where $\bbf = \zerobf$ and the resulting constrained lasso solution can be translated back to the original generalized lasso parameterization via an affine transformation, in line with the result in this section that this is a situation where the constrained lasso can in fact be transformed to a generalized lasso.

Now consider the more general case with an arbitrary $\bbf \ne \zerobf$.  We can re-arrange the equality constraints  as 
\begin{eqnarray}
	\Abf \betabf = \bbf 	\nonumber \\
	\Abf \betabf - \bbf = \zerobf \nonumber \\
	\begin{pmatrix} \Abf,  -\Ibf_q \end{pmatrix} \begin{pmatrix} \betabf \\ \bbf \end{pmatrix} = \tilde{\Abf} \tilde{\betabf} = \zerobf,	\nonumber
\end{eqnarray}
 and then apply the above result using  $\tilde{\Dbf} = \tilde{\Qbf}_2$ from the QR decomposition of $\tilde{\Abf}^T$.  However, now the reparameterized problem is
 \begin{eqnarray}
	\tilde{\betabf} = \tilde{\Dbf} \tilde{\thetabf} \begin{pmatrix} \tilde{\Dbf}_{1} \tilde{\thetabf}_1 \\ \tilde{\Dbf}_{2} \tilde{\thetabf}_2 \end{pmatrix} = \begin{pmatrix} \betabf \\ \bbf \end{pmatrix} \nonumber,
 \end{eqnarray}
 we are still left with the constraint $\tilde{\Dbf}_{2} \tilde{\thetabf}_2 = \bbf$.  Therefore, for equality constraints with $\bbf \ne \zerobf$, a constrained lasso can be transformed into a constrained generalized lasso.  This result is trivial, however, since a constrained lasso is always a constrained generalized lasso with $\Dbf = \Ibf$.

\subsubsection{Null-Space Method}
Another common method for solving least squares problems with equality constraints (LSE) is the null-space method \citep{bjorck2015numerical}.  To apply this method to the constrained lasso, we again restrict our attention to a constrained lasso with only equality constraints for the time being,
\begin{eqnarray}
	&\text{minimize}& \quad \frac 12 \| \ybf - \Xbf \betabf\|_2^2 + \rho \|\betabf\|_1 \label{eqn:LS-constrlassoEqb} \\
	&\text{subject to}& \quad \Abf \betabf = \bbf.	\nonumber
\end{eqnarray}
We will show that the application of the null-space method results in a shifted generalized lasso.  Consider the QR decomposition of $\Abf^T \in \reals^{p \times q}$, where $\text{rank}(\Abf) = q$, 
\begin{eqnarray*}
	\Abf^T = \Qbf \begin{pmatrix} \Rbf \\ \zerobf \end{pmatrix} \text{ with } \Qbf\Qbf' = \Ibf,
\end{eqnarray*}
and $\Qbf \in \reals^{p \times p}$, $\Rbf \in \reals^{q \times q}$, and $\zerobf \in \reals^{p-q \times q}$.  Let 
\begin{eqnarray*}
	\Xbf \Qbf = \begin{pmatrix} \Xbf_1,  \Xbf_2 \end{pmatrix} \text{ and } \Qbf'\betabf = \begin{pmatrix} \alphabf_1 \\ \alphabf_2 \end{pmatrix},
\end{eqnarray*}
with $\Xbf_1 \in \reals^{n \times q}$, $\Xbf_2 \in \reals^{n \times p-q}$, $\alphabf_1 \in \reals^{q \times 1}$, and $\alphabf_2 \in \reals^{p-q \times 1}$, then 
\begin{eqnarray*}
	\Xbf\betabf = \Xbf \Qbf \Qbf^T \betabf =  \begin{pmatrix} \Xbf_1,  \Xbf_2 \end{pmatrix} \begin{pmatrix} \alphabf_1 \\ \alphabf_2 \end{pmatrix} =  \Xbf_1\alphabf_1 + \Xbf_2\alphabf_2.
\end{eqnarray*}
As for the constraints, we have 
\begin{eqnarray*}
	\Abf = \begin{pmatrix} \Rbf^T ,  \zerobf^T \end{pmatrix} \Qbf^T \Rightarrow \Abf\beta = \begin{pmatrix} \Rbf^T ,  \zerobf^T \end{pmatrix} \Qbf^T \betabf = \begin{pmatrix} \Rbf^T ,  \zerobf^T \end{pmatrix} \begin{pmatrix} \alphabf_1 \\ \alphabf_2 \end{pmatrix} =  \Rbf^T \alphabf_1.
\end{eqnarray*}
Lastly, the penalty term becomes 
\begin{eqnarray*}
	\|\betabf\|_1 = \| \Qbf \begin{pmatrix} \alphabf_1 \\ \alphabf_2 \end{pmatrix} \|_1 = \| \Qbf_1 \alphabf_1 + \Qbf_2 \alphabf_2 \|_1.
\end{eqnarray*}
Thus, putting these pieces together we can re-write \eqref{eqn:LS-constrlassoEqb} as
\begin{eqnarray}
	&\text{minimize}& \quad \frac 12 \| (\ybf - \Xbf_1 \alphabf_1) - \Xbf_2 \alphabf_2 \|_2^2 + \rho  \| \Qbf_1 \alphabf_1 + \Qbf_2 \alphabf_2 \|_1 \label{eqn:LS-constrlassoEqTrans} \\
	&\text{subject to}& \quad \Rbf^T \alphabf_1 = \bbf.	\nonumber
\end{eqnarray}
Since $\Rbf^T$ is invertible, we can further directly incorporate the constraints in the objective function by plugging in $\alphabf_1 = (\Rbf^T)^{-1}\bbf$, 
\begin{eqnarray}
	&\text{minimize}& \quad \frac 12 \| (\ybf - \Xbf_1 (\Rbf^T)^{-1}\bbf) - \Xbf_2 \alphabf_2 \|_2^2 + \rho  \| \Qbf_1 (\Rbf^T)^{-1}\bbf + \Qbf_2 \alphabf_2 \|_1 \label{eqn:LS-constrlassoEqTrans2} ,
\end{eqnarray}
or more concisely as
\begin{eqnarray}
	&\text{minimize}& \quad \frac 12 \| \tilde{\ybf} - \Xbf_2 \alphabf_2 \|_2^2 + \rho  \| \cbf + \Qbf_2 \alphabf_2 \|_1 \label{eqn:LS-constrlassoEqTrans3} ,
\end{eqnarray}
with  $\tilde{\ybf} = \ybf - \Xbf_1 (\Rbf^T)^{-1}\bbf$ and $\cbf = \Qbf_1 (\Rbf^T)^{-1}\bbf$.  So \eqref{eqn:LS-constrlassoEqTrans3} resembles an unconstrained generalized lasso problem with $\Dbf = \Qbf_2$, but it has been shifted by a constant vector $\cbf = \Qbf_1 (\Rbf^T)^{-1}\bbf$ that can not be decoupled from the penalty term.  Therefore, it once again is not possible to solve a constrained lasso by reformulating it as an unconstrained generalized lasso.  It should be noted that such a transformation is not possible even in the presence of only equality constraints, and the addition of inequality constraints would only further complicate matters.  As pointed out by \citet{gentle07}, there is no general closed-form solution to least squares problems with inequality constraints.

\subsection{Subgradient Violations}
\label{sec:subgrad}
As pointed out in Section~\ref{sec:path}, there is the potential for the subgradient conditions \eqref{eq:subgrad} to become violated if an inactive coefficient is moving too slowly.  Here we provide the supporting derivations for this result and what is meant by too slowly.  To preview the result, an inactive coefficient, $j \in {\cal A}^c$, with subgradient $s_j = \pm 1$ is  moved to the active set if $s_j \cdot \frac{d}{d \rho} [\rho s_j ] < 1$.  To see this, without loss of generality, assume that $s_j = -1$ for some inactive coefficient $\beta_j$, $j \in {\cal A}^c$.  As given in Table~\ref{tab:events}, since $\rho$ is decreasing, $s_j$ is updated along the path via
			\begin{eqnarray*}
	[\rho^{(t+1)} s_j^{(t+1)}] =  \rho^{(t)} s_j^{(t)} - \Delta \rho \cdot\frac{d}{d \rho} [\rho s_j],
			\end{eqnarray*}
which implies 
			\begin{eqnarray}
	s_j^{(t+1)} = \left( \rho^{(t)} s_j^{(t)} - \Delta \rho \cdot\frac{d}{d \rho} [\rho s_j] \right)/\rho^{(t+1)}. \label{eqn:subgradTrouble1}
			\end{eqnarray}
Since $\rho$ is decreasing, $\rho^{(t)} > \rho^{(t+1)}$,  but we define $\Delta \rho > 0$ which implies that $\Delta \rho = \rho^{(t)} - \rho^{(t+1)}$.  Using this and $s_{j}^{(t)} = -1$, then for a given inactive coefficient $j \in {\cal A}^c$, \eqref{eqn:subgradTrouble1} becomes
 		\begin{eqnarray}
		s_{j}^{(t+1)} &=& \left( -\rho^{(t)} - (\rho^{(t)} - \rho^{(t+1)}) \frac{d}{d \rho} [\rho s_{j}] \right)/\rho^{(t + 1)}. \label{eqn:subgradTrouble2}
		\end{eqnarray}
To identify the trouble ranges for $\frac{d}{d \rho} [\rho s_{j}]$ that would result in a violation of the subgradient conditions, we can rearrange \eqref{eqn:subgradTrouble2} as follows,
 		\begin{eqnarray}
		s_{j}^{(t+1)} &=& \left( -\rho^{(t)} - (\rho^{(t)} - \rho^{(t+1)}) \frac{d}{d \rho} [\rho s_{j}] \right)/\rho^{(t+1)} \nonumber \\
			&=& -\frac{d}{d \rho} [\rho s_{j}] \left( \frac{\rho^{(t)}}{\rho^{(t + 1)}} - 1 \right) - \frac{\rho^{(t)}}{\rho^{(t+1)}} \nonumber \\
			&=& -\frac{d}{d \rho} [\rho s_{j}] \left( \frac{\rho^{(t)}}{\rho^{(t+1)}} - 1 \right) - \frac{\rho^{(t)}}{\rho^{(t+1)}} + 1 - 1 \nonumber \\
			&=& - \frac{d}{d \rho} [\rho s_{j}] \left( \frac{\rho^{(t)}}{\rho^{(t+1)}} - 1 \right) + \left(1 - \frac{\rho^{(t)}}{\rho^{(t+1)}} \right) - 1 \nonumber \\
			&=& \left( \frac{d}{d \rho} [\rho s_{j}] + 1 \right) \left(1 -  \frac{\rho^{(t)}}{\rho^{(t+1)}} \right) - 1. \label{eqn:subgradTrouble3}
		\end{eqnarray}
The second term in the product in \eqref{eqn:subgradTrouble3} is always negative, since $\rho^{(t)} > \rho^{(t+1)} \Rightarrow \displaystyle \frac{\rho^{(t)}}{\rho^{(t+1)}} > 1 \Rightarrow 0 > 1 - \frac{\rho^{(t)}}{\rho^{(t+1)}}$.  Now, consider different values for $\frac{d}{d \rho} [\rho s_{j}]$:

\begin{enumerate}[i)]
	\item $\frac{d}{d \rho} [\rho s_{j}] > -1$:  When $\frac{d}{d \rho} [\rho s_{j}] > -1$, then $ \left( \frac{d}{d \rho} [\rho s_{j}] + 1 \right) > 0$,  so the product term in \eqref{eqn:subgradTrouble3} involves a positive number multiplied by a negative number and is thus negative.  However, this would lead to $s_j < -1$ when 1 is subtracted from the product term, which is a violation of the subgradient conditions.  
	\item $\frac{d}{d \rho} [\rho s_{j}] = -1$: This is fine as it maintains $s_j = -1$, since $\frac{d}{d \rho} [\rho s_{j}] = -1 \Rightarrow \left( \frac{d}{d \rho} [\rho s_{j}] + 1 \right) = 0 \Rightarrow s_{j}^{(t+1)} = -1$.
	\item $\frac{d}{d \rho} [\rho s_{j}] < -1$:  This situation is also fine as $\frac{d}{d \rho} [\rho s_{j}] < -1 \Rightarrow \left( \frac{d}{d \rho} [\rho s_{j}] + 1 \right) < 0$, so the product term in \eqref{eqn:subgradTrouble3} is positive and the subgradient is moving towards zero, which is fine since $j \in {\cal A}^c$.  
\end{enumerate}
The only issue, then, arises when $\frac{d}{d \rho} [\rho s_{j}] > -1$.  The corresponding range for $j \in {\cal A}^c$ but $s_j = 1$ is $\frac{d}{d \rho} [\rho s_{j}] < -1$, which is derived similarly.  Combining these two situations, the range to monitor can be written more succinctly as $s_j \cdot \frac{d}{d \rho} [\rho s_j] < 1$.  Thus, to summarize, an inactive coefficient $j \in {\cal A}^c$ with subgradient $s_j = \pm 1$ and $s_j \cdot \frac{d}{d \rho} [\rho s_j] < 1$ needs to be moved back into the active set, ${\cal A}$, before the path algorithm proceeds to prevent a violation of the subgradient conditions \eqref{eq:subgrad}.

\end{appendix}

% Brian's bib
\bibliography{brianBib}
\bibliographystyle{asa}

\end{document}